\documentclass[11pt,a4paper]{article}
\usepackage[T1]{fontenc}
\usepackage[utf8]{inputenc}
\usepackage{libertinus}
\usepackage[scaled=0.86]{beramono}
\usepackage[margin=2.5cm]{geometry}
\usepackage{microtype}
\usepackage{amsmath,amssymb}
\usepackage{booktabs}
\usepackage{threeparttable}
\usepackage{tabularx}
\usepackage{makecell}
\newcolumntype{Y}{>{\raggedright\arraybackslash}X}
\usepackage{xcolor}
\usepackage{graphicx}
\usepackage{tikz}
\usetikzlibrary{arrows.meta,positioning,decorations.pathreplacing,calc,patterns}
\usepackage{pgfplots}
\pgfplotsset{compat=1.18}
\usepgfplotslibrary{groupplots,fillbetween}
\usepackage[numbers,sort&compress]{natbib}
\usepackage[hidelinks]{hyperref}
\usepackage{enumitem}
\setlist{itemsep=2pt,topsep=4pt}

\definecolor{navy}{HTML}{0B1F3A}
\definecolor{accent}{HTML}{C8102E}
\definecolor{covc}{HTML}{2E6DB4}
\definecolor{histc}{HTML}{8A8F98}
\newcommand{\m}[1]{\texttt{#1}}

\title{\textbf{EnergyBench: a live, gate-aligned benchmark\\of day-ahead forecasts for European power markets}}
\author{Evgenii Genov$^{1}$ \qquad Ander Martinez Alonso$^{1,2}$ \qquad Christoph Bergmeir$^{3,4}$\\[4pt]
\small $^{1}$EVERGi Research Group, Flanders Alliance for Climate Technology (FACT),\\
\small Vrije Universiteit Brussel, Pleinlaan 2, 1050 Brussels, Belgium\\
\small $^{2}$DTU Wind and Energy Systems, Technical University of Denmark\\
\small $^{3}$Department of Computer Science and Artificial Intelligence, University of Granada, Granada, Spain\\
\small $^{4}$Department of Data Science \& Artificial Intelligence, Monash University,\\
\small 25 Exhibition Walk, Clayton, VIC 3168, Australia\\[4pt]
\small\href{mailto:evgenii.genov@vub.be}{evgenii.genov@vub.be}, \href{mailto:Ander.Martinez.Alonso@vub.be}{Ander.Martinez.Alonso@vub.be},
\href{mailto:bergmeir@ugr.es}{bergmeir@ugr.es}}
\date{\small Methodology preprint, November 2026\\
\href{https://energybench.foretec.co}{energybench.foretec.co} \quad\textperiodcentered\quad
\href{https://github.com/ujohn33/foretec-energy-bench}{github.com/ujohn33/foretec-energy-bench}}

\begin{document}
\maketitle

\begin{abstract}
\noindent
Energy forecasting models are usually compared on backtests and fixed public datasets. Backtests can use data that
did not yet exist when the forecast would have been made, and a fixed test set, reused across studies and possibly
contained in the pretraining data of foundation models, invites overfitting to the benchmark. EnergyBench evaluates
every entry live instead. Each day at 11:30 CET/CEST, thirty minutes before the day-ahead (SDAC) gate closes,
every model forecasts the next day's price, wind and solar generation and load on the 15-minute grid for six bidding
zones (BE, NL, FR, DE-LU, DK1, DK2). Unlike existing live benchmarks, EnergyBench runs the models itself, on one
information set that it freezes at the cut-off and publishes, including three numerical weather prediction models, so that
foundation models can be compared with and without the same weather covariates under the actual market deadline. Errors are
pooled per series against a seasonal-naive baseline, and models are ranked by a Bradley--Terry rating on the Elo scale
with a day-bootstrap interval. The field spans baselines, linear and gradient-boosted models on the frozen weather,
eighteen time-series foundation models from eleven families, ensembles, and the grid operators' own forecasts. Over the
first thirty-two live days, the covariates reduced the error of each of the ten foundation models run with and without
them by 37--51\% on matched races (64--73\% on wind, 48--61\% on solar). Used zero-shot, these models match or outperform a LightGBM model trained on the same covariates. Model size,
from 3M to 331M parameters, has a much smaller effect, and median ensembles of the covariate models rank highest. The grid
operators' pre-gate forecasts remain ahead on wind and solar, where the best single foundation models have 16--22\% more error; on load the gap is 4--6\%, within the intervals.
\end{abstract}

\section{Introduction}

Day-ahead forecasts of price, renewable generation and load are central inputs to bidding decisions on European
power markets, yet reported accuracies of different models are rarely comparable~\cite{lago,hewamalage}. Three problems
commonly affect backtests~\cite{hewamalage}. First, the information set drifts: a forecast nominally made at 11:30 is built from weather runs that
were published hours later, from actuals that were still provisional, or from a TSO forecast that only appears after
the gate~\cite{energyarena}. Second, the target may differ from the quantity the market trades, for example hourly values where the
market now trades quarter-hours, generation definitions that differ between data providers, or actuals that were later
revised. Third, the evaluation window, the baseline and the metric can be chosen after the results are known. Fixed
public datasets add a related problem: when the same test set is reused across studies, models and settings are
selected on it repeatedly, and for foundation models the test period may also be part of the pretraining
data~\cite{meyerleak,impermanent}.

A fourth problem concerns the evaluation of time-series foundation models. In operational energy forecasting, exogenous inputs are
the standard, not an option: wind and solar forecasts are built on numerical weather predictions, load forecasts on
temperature and calendar effects, and day-ahead price forecasts on forecasts of load, renewable infeed and fuel
cost~\cite{hongreview,weron,lago}. The grid operators' own forecasts, the industry reference, are NWP-driven, and the
energy forecasting competitions routinely supplied weather forecasts~\cite{gefcom}. The benchmarks on which
foundation models are compared are, by contrast, mostly univariate; fev-bench~\cite{fevbench} and, for energy,
FETS~\cite{fets} are recent exceptions. A model that forecasts tomorrow's solar output from
the last four weeks of solar output alone is evaluated on a task that does not arise in practice, and a ranking
obtained in this way is of limited use for choosing a model to deploy.

A live benchmark that provides every entry with the same inputs avoids the first two problems and the risk of contamination. If each
model runs at the actual cut-off on the data available at that time, weather forecasts included, and its forecast is
locked before the actual gate, look-ahead is excluded. Because the outcomes do not yet exist when the forecasts are
made, no model can have been trained or tuned on them, and the record grows daily and is public. The drawback is
that a live record is short and its standings vary from day to day. The design described below accounts for this
variability. Errors are pooled before they are normalised, rankings are reported with intervals, and only live
forecasts are reported. The third problem, the choice of analysis after the results are known, is reduced but not
removed by fixed public rules (Section~\ref{sec:limits}).

EnergyBench began operating on 2~October 2026 with three zones (BE, NL, FR) and three targets; total load was added on
6~October and three zones (DE-LU, DK1, DK2) on 15~October.
The code and the configuration file that define the benchmark are versioned in a public repository. This paper
describes that configuration as of 31~October 2026 and the live results up to issue day 2~November.

\section{Related work}

\paragraph{Static benchmarks.} Forecasting competitions have shaped the field since the M-competitions~\cite{m4}; the
Monash archive~\cite{monash}, GIFT-Eval~\cite{gifteval} and fev-bench~\cite{fevbench} standardise the evaluation of
general-purpose and foundation time-series models on fixed datasets; 46 of fev-bench's 100 tasks include
covariates. In energy, the Global Energy Forecasting
Competitions~\cite{gefcom} posed realistic tasks as one-off events, and Lago et al.~\cite{lago} published an open
benchmark for day-ahead price forecasting with fixed splits and day-ahead exogenous forecasts for five markets.
FETS~\cite{fets} compares foundation models, with and without covariates, against task-specific gradient-boosted and
random-forest models on 54 public energy datasets in nine categories, with archived ECMWF IFS forecasts as weather
covariates over the forecast horizon, and finds covariate-informed zero-shot foundation models best in aggregate. Fixed
datasets have two weaknesses for the present purpose. They describe a past regime, and energy systems change quickly;
and their test periods may have been seen, directly or through pretraining corpora, by the models they evaluate.
Meyer et al.~\cite{meyerleak} document both direct overlap, where one model's training data serve as another's test
set, and temporal overlap between correlated training and test series in the evaluation of foundation models.
Retrospective energy studies also often leave the information set implicit: several price-forecasting studies use
ENTSO-E's day-ahead wind and solar forecasts as inputs although those are published only after 18:00 on $D-1$, six
hours after the auction closes~\cite{energyarena}.

\paragraph{Covariates.} The value of weather inputs in energy forecasting is long established. Wind and solar power
forecasting has been NWP-based since its beginnings, and the wind and solar tracks of GEFCom2014 gave every team ECMWF
forecasts~\cite{gefcom}; load forecasting rests on temperature, and the reviews of price forecasting single out
forecasts of load and renewable generation as the most informative exogenous drivers~\cite{weron,lago,hongreview}.
Foundation models have only recently gained the means to use such inputs: Chronos-2, TimesFM, TiRex-2, Timestral, TimeH2O and
the t0 models accept known-future covariates, by different mechanisms and with different success. Whether and how much they
benefit cannot be measured on a univariate benchmark.

\paragraph{Live benchmarks.} Three recent benchmarks score forecasts made before the outcome is known; the two that
cover energy are compared with EnergyBench in Table~\ref{tab:compare}. TS-Arena~\cite{tsarena} pre-registers forecasts
of time-series foundation models for German, US and Finnish price, load and generation series and ranks them by an
online Elo rating. It provides no covariates, so the models' use of weather forecasts, which in our results reduces
their error by 37--51\% (Section~\ref{sec:results}), is not assessed; and each round starts from a series' last
published value rather than at a market gate, so the forecasts do not correspond to bids that could have been placed.
Energy-Arena~\cite{energyarena} is gate-aligned: forecasts for the next day are submitted before the 12:00 day-ahead
gate. It compares submissions rather than models under common conditions, since participants run their own models on
inputs of their choice, which can be neither verified nor reproduced. Impermanent~\cite{impermanent}, from the
TimeCopilot project, runs the models itself and scores them as the observations arrive, motivated by the contamination
of static test sets and by model selection on them; its series are univariate counts of GitHub activity rather than
energy series. In industry, the Predico platform~\cite{predico} scores live renewable forecasts submitted to Elia and
50Hertz against the real gate, for enrolled forecasters and without publishing their inputs.

\paragraph{Positioning.} EnergyBench takes Energy-Arena's operational framing (the day-ahead gate defines the task)
and TS-Arena's pre-registration logic (only a forecast made before the outcome is known is free of leakage), and, like
Impermanent, runs every model itself, on an information set that includes weather forecasts. Each day it fetches one information set at the gate-aligned
cut-off, freezes and publishes it, and gives it to every entry: the information set is enforced rather than declared,
and any forecast can be reproduced from its archived inputs. Foundation models from seven families also run with the
same frozen covariates, so the effect of the weather inputs is measured for each model; all four targets
are forecast in six zones; and the grid operators' own forecasts, before and after the gate, serve as references. In
return, entries must be submitted as code rather than as forecasts, and proprietary data cannot be used.
Table~\ref{tab:compare} summarises the differences.

\begin{table}[t]
\centering\small
\caption{Live energy forecasting benchmarks compared, as described in their papers and on their websites (October 2026).}
\label{tab:compare}
\begin{tabularx}{\linewidth}{@{}>{\raggedright\arraybackslash}p{2.7cm}YYY@{}}
\toprule
& TS-Arena~\cite{tsarena} & Energy-Arena~\cite{energyarena} & EnergyBench (this paper) \\
\midrule
Task timing & from each series' last published value; no market gate & submission before the 12:00 day-ahead gate, whole next day & 11:30 cut-off, locked before the 12:00 gate, whole next day \\
Zones and targets & 14 challenges (SMARD, Gridstatus, Fingrid) & wind, solar, load: AT, DE-LU; price: 15 zones & price, wind, solar, load in each of BE, NL, FR, DE-LU, DK1, DK2 \\
Who runs models & participants (containers) & participants (API submissions) & the benchmark, from submitted code \\
Information set & series history only & anything before the gate; self-declared, unverified & frozen and published: history, three NWP models, load forecast, fuel \\
Covariates for foundation models & none & if a participant adds them & eleven foundation models also with the same covariates \\
Foundation models & the focus of the benchmark & none named; depends on submissions & eleven families, and ensembles \\
Transparency & platform code public & method description and forecasts optional & code, inputs and forecasts public \\
Scores & MASE & RMSE, WIS, CRPS & MAE, RMSE, bias, pinball; relative to seasonal naive, pooled per series \\
Ranking & online Elo, replay bootstrap & per metric over rolling windows & Bradley--Terry Elo, day bootstrap \\
References & statistical baselines & optional benchmark models & baselines; the TSOs' forecasts before and after the gate \\
\bottomrule
\end{tabularx}
\end{table}

\section{The benchmark}

This section defines the benchmark as it runs: the forecasting task, the daily timeline and what a model may know at
the cut-off, where the actuals come from, the covariates frozen for every entry, the field of entries, and how
forecasts are scored, ranked and valued.

\subsection{Task}

For each zone $z \in \{\text{BE}, \text{NL}, \text{FR}, \text{DE-LU}, \text{DK1}, \text{DK2}\}$ and target $y \in
\{\text{price}, \text{wind}, \text{solar}, \text{load}\}$ (24 series), a forecast made on issue day $D-1$ covers
every quarter-hour of delivery day $D$ in local time: 96 values, or 92 and 100 on clock-change days. Price is the SDAC
day-ahead price (EUR/MWh); wind is onshore plus offshore generation, solar is solar generation and load is total load
(MW). Models return a point forecast and optionally the nine deciles $q_{0.1},\dots,q_{0.9}$.

\subsection{Timeline and information set}

\begin{figure}[t]
\centering
\begin{tikzpicture}[x=1cm,y=1cm,font=\scriptsize]
  % piecewise time axis: D-1 06:00..24:00 at 0.5 cm/h, day D at 0.125 cm/h, D+1 00:00..16:00 at 0.25 cm/h
  \fill[covc!12] (9,0.08) rectangle (12,0.5); \node[covc] at (10.5,0.29) {delivery day $D$};
  \draw[-{Latex[length=2mm]},thick,navy] (0,0) -- (16,0);
  \foreach \x/\l in {3/{$D{-}1$ 12:00},9/{$D$ 00:00},12/{$D{+}1$ 00:00},15/{$D{+}1$ 12:00}}
    {\draw[navy] (\x,-0.1) -- (\x,0.1); \node[below=2pt,navy] at (\x,-0.05) {\l};}
  \newcommand{\ev}[5]{\draw[#3,thick] (#1,0) -- (#1,#2); \node[anchor=#4,#3,inner sep=1.5pt] at (#1,#2) {#5};}
  \ev{2.75}{2.15}{navy}{south west}{\textbf{11:30 cut-off}: snapshot, frozen NWP runs, every model runs}
  \ev{3.0}{1.45}{accent}{south west}{\textbf{12:00 gate}: a run finishing later is never scored}
  \ev{4.375}{0.75}{navy}{south west}{prices scored at 14:45}
  \ev{5.5}{-0.75}{histc}{north west}{16:15--18:00 TSO final forecasts (\m{tso\_final})}
  \ev{7.375}{1.1}{navy}{south west}{late prices scored at 20:45}
  \ev{13.75}{0.75}{navy}{south east}{wind, solar, load scored at 07:00 on $D{+}1$}
  \draw[decorate,decoration={brace,mirror,amplitude=3pt},histc] (13.75,-0.6) -- (16,-0.6)
    node[midway,below=3pt,anchor=north east,xshift=1.1cm,histc] {$D{+}1\dots D{+}3$: revision checks};
\end{tikzpicture}
\caption{One benchmark day (not to scale). Everything a model sees is fixed at the 11:30 cut-off; the forecast must
be locked before the 12:00 gate. The TSOs' final day-ahead wind and solar forecasts appear only after the gate and
enter as references.}
\label{fig:timeline}
\end{figure}
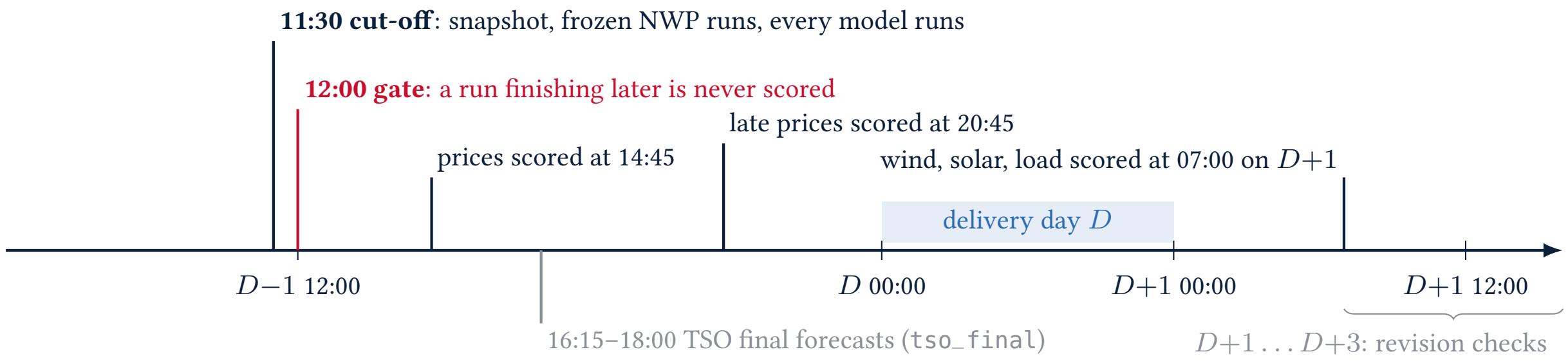

The benchmark day is defined by three clock times in CET/CEST (Figure~\ref{fig:timeline}). At the \emph{cut-off},
11:30 on $D-1$, a snapshot of everything public is taken and every model runs on it. The \emph{gate} is 12:00, the
SDAC gate closure. A live run that finishes after the gate is marked late and never scored, and a day's forecasts
cannot be changed once written, so the record consists of the forecasts as they were at the gate. Prices are scored on
$D-1$ after the auction for $D$ has cleared, at the scoring run of 14:45, or at that of 20:45 if the prices were not yet
published. Wind, solar and load are scored at 07:00 on $D+1$ if every quarter-hour of $D$ has been published, and
otherwise at the first later scoring run at which 95\% has.

What a model may know at the cut-off is set per target. Prices: everything up to the end of $D-1$, which the auction
on $D-2$ published. Wind, solar and load: actuals up to the cut-off minus one hour of publication lag. Every model
receives the last 28 days of 15-minute history (the context). A model may train on longer history, provided that each
training row uses only data from before that row's own cut-off. Two qualifications apply: the training actuals are the
current values, which may have been revised since they were first published, and the weather in the two-year training
frames of two entries is matched to the live lead time rather than to the live run (Section~\ref{sec:cov}). Training
never takes place inside the live window. Models that retrain do so in the nightly catch-up at 03:30. The TSOs' day-ahead renewable forecasts are not
an input: Regulation (EU) 543/2013~\cite{reg543} has them published by 18:00 on $D-1$, after the gate.

\subsection{Actuals and provenance}
\label{sec:actuals}

The primary source is the ENTSO-E Transparency Platform~\cite{entsoe}, parsed from its XML: day-ahead prices (document
A44, the 15-minute SDAC prices), actual generation per production type (A75; wind as onshore B19 plus offshore B18,
solar B16) and actual total load (A65). ENTSO-E does not deliver every quarter-hour: alone it lacked 12\% of Belgian
and 2.6\% of French wind quarter-hours in September 2026. Gaps are filled from the TSOs' own open data (Elia for
Belgium, RTE éCO2mix for France), then from Energy-Charts. Where a TSO publishes finer data than it sends ENTSO-E, or a
quantity closer to the national total, the TSO is the primary source (Table~\ref{tab:sources}). The source of every
value is recorded.\footnote{The sources for DE-LU, DK1 and DK2 follow those of Electricity Maps~\cite{electricitymaps},
except for Danish wind: ENTSO-E's DK1 offshore series covers only the parks registered there and has missed
0.3--0.5\,GW of offshore capacity since July 2026, so Danish wind has come from Energinet since delivery day 18
October. The first two issue days, 2 and 3 October, were scored on Energy-Charts data alone.}

\begin{table}[t]
\centering\small
\caption{Primary source of the actuals per zone and target. Gaps are filled from the TSO's own data (Elia, RTE
éCO2mix), then from Energy-Charts; the Dutch NED series have no fallback.}
\label{tab:sources}
\begin{tabular}{@{}lllll@{}}
\toprule
& Price & Wind & Solar & Load \\
\midrule
ENTSO-E document & A44 & A75 (B18 + B19) & A75 (B16) & A65 \\
\midrule
BE & ENTSO-E & Elia (quarter-hourly) & Elia (quarter-hourly) & ENTSO-E \\
NL & ENTSO-E & NED (national) & NED (national) & NED (gross) \\
FR & ENTSO-E & ENTSO-E & ENTSO-E & ENTSO-E \\
DE-LU & ENTSO-E & ENTSO-E & ENTSO-E & ENTSO-E \\
DK1, DK2 & ENTSO-E & Energinet & ENTSO-E & ENTSO-E \\
\bottomrule
\end{tabular}
\end{table}

In the Netherlands, ENTSO-E's series cover only what TenneT meters itself: about 1\% of the solar fleet (a capacity
factor of 0.1\% over 20 September--7 October 2026, against 10.7\% in the national estimate), mostly offshore wind, and
load net of embedded solar (about 6\,GW against 17\,GW gross at a sunny midday). The Dutch wind, solar and load
targets therefore come from NED, the Nationaal Energie Dashboard of TenneT and the distribution operators, which
publishes national estimates: wind as onshore plus metered offshore, load as gross consumption as Elia and RTE report
it. These series have no fallback, since filling a gap from ENTSO-E would combine two definitions within one delivery
day; quarter-hours that NED has not published are left unscored.\footnote{Dutch forecasts made against the earlier
ENTSO-E definition (issue days up to 10 October for wind and solar, 11 October for load) are not scored. NED's offshore
wind was a park-by-park model through delivery day 12 October, 7--27\% above the metered output used since; those days
keep their scores.}

Actuals are \emph{frozen} the first time they are complete and scored from the frozen copy. Until $D+3$, wind, solar
and load are re-fetched at every scoring run; a change above 0.5\% of the day's energy re-freezes the
actuals and re-scores the day, and every check is logged. Prices are final at the auction. Between the gate and
scoring, the site shows the quarter-hours metered so far as provisional actuals, which are never scored.

\subsection{Covariates frozen at the cut-off}
\label{sec:cov}

Models that declare covariate inputs receive, per issue day, a table frozen at the cut-off and a log of what went
into it.

\begin{figure}[t]
\centering
\begin{tikzpicture}[x=0.95cm,y=1cm,font=\scriptsize]
  \foreach \yy/\cut/\gate/\lab in {0/9.5/10/{Summer time (CEST, UTC+2): cut-off 09:30 UTC, gate 10:00 UTC},-3.3/10.5/11/{Winter time (CET, UTC+1): cut-off 10:30 UTC, gate 11:00 UTC}} {
    \begin{scope}[yshift=\yy cm]
      \node[anchor=west,font=\scriptsize\bfseries] at (0,1.2) {\lab};
      \node[anchor=east] at (-0.7,0.15) {ICON-EU};
      \node[anchor=east] at (-0.7,-0.75) {GFS, IFS};
      \foreach \i/\y/\nm in {0/0.45/00,3/0.25/03,6/0.05/06,9/-0.15/09} {
        \pgfmathsetmacro{\av}{\i+3.5}
        \pgfmathtruncatemacro{\ok}{and(\av<=\cut+0.001, \av+3>\cut+0.001) ? 1 : 0}
        \ifnum\ok=1 \draw[covc,line width=2.2pt] (\i,\y) -- (\av,\y); \fill[covc] (\i,\y) circle (2pt);
        \else \draw[gray!40,line width=2.2pt] (\i,\y) -- (\av,\y); \fill[gray!60] (\i,\y) circle (1.6pt); \fi
        \node[anchor=east,gray,inner sep=1.5pt] at (\i-0.05,\y) {\nm Z};}
      \foreach \i/\y/\nm in {0/-0.65/00,6/-0.85/06} {
        \pgfmathsetmacro{\av}{\i+7}
        \pgfmathtruncatemacro{\ok}{and(\av<=\cut+0.001, \av+6>\cut+0.001) ? 1 : 0}
        \ifnum\ok=1 \draw[covc,line width=2.2pt] (\i,\y) -- (\av,\y); \fill[covc] (\i,\y) circle (2pt);
        \else \draw[gray!40,line width=2.2pt] (\i,\y) -- (\av,\y); \fill[gray!60] (\i,\y) circle (1.6pt); \fi
        \node[anchor=east,gray,inner sep=1.5pt] at (\i-0.05,\y) {\nm Z};}
      \draw[gray!60] (0,-1.2) -- (14,-1.2);
      \foreach \h in {0,2,...,14} {\draw[gray!60] (\h,-1.2) -- ++(0,-0.08) node[below,gray,inner sep=1pt] {\h:00};}
      \draw[navy,thick] (\cut,0.65) -- (\cut,-1.2) node[pos=0,above left=-1pt,navy] {cut-off};
      \draw[accent,thick,dashed] (\gate,0.65) -- (\gate,-1.2) node[pos=0,above right=-1pt,accent] {gate};
    \end{scope}}
\end{tikzpicture}
\caption{NWP runs admitted at the cut-off, in UTC. Each bar runs from a run's initialisation (dot) to the time it is
admitted, initialisation plus the lag (3.5\,h for ICON-EU, 7\,h for GFS and IFS); blue: the run the covariates use. The
cut-off stays at 11:30 local time, so it moves by an hour in UTC with the clock change, but the same runs are admitted
in both regimes: ICON-EU 06 UTC (exactly at the summer cut-off) and GFS and IFS 00 UTC.}
\label{fig:nwp}
\end{figure}
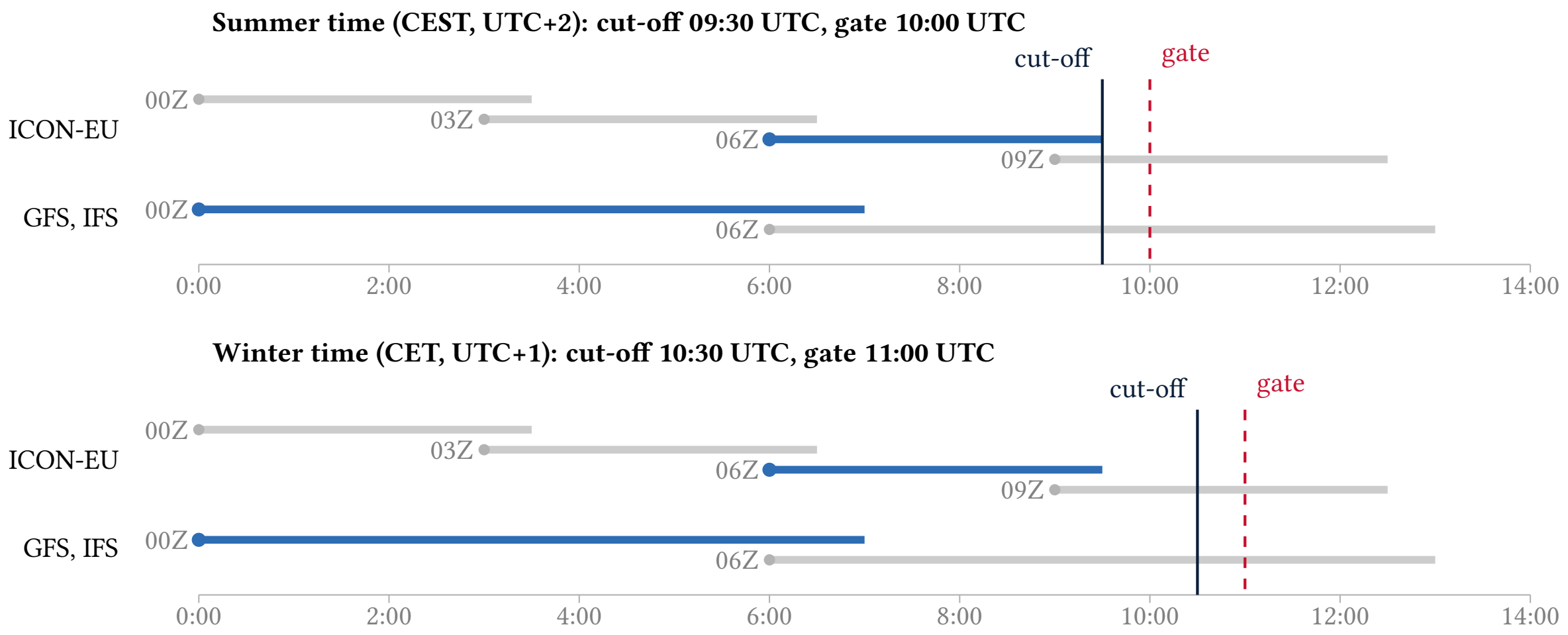

\begin{figure}[t]
\centering
\includegraphics[width=0.72\linewidth]{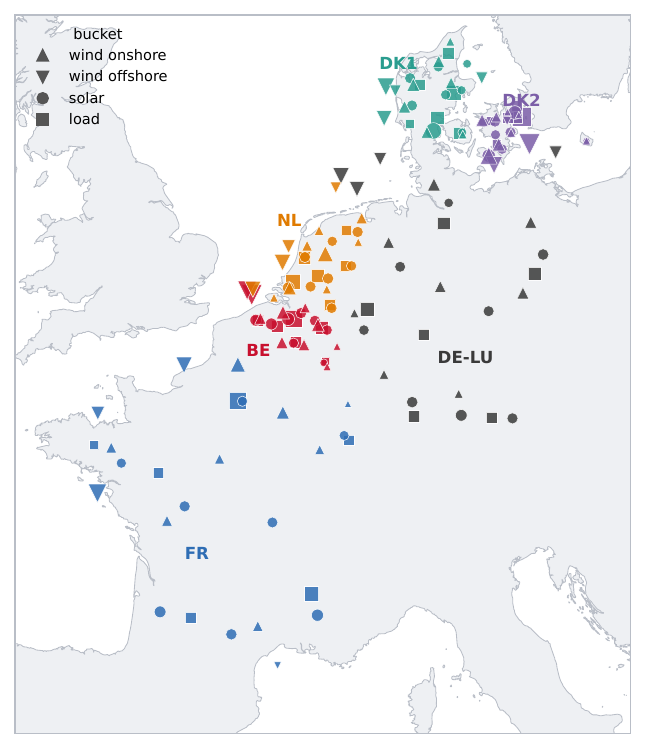}
\caption{The 153 weather sampling points: capacity-weighted $k$-means centroids per zone and bucket, from public
registers (load: population). Marker size is the point's share of its zone and bucket.}
\label{fig:map}
\end{figure}

\paragraph{Weather.} Three independent global and regional NWP models, ICON-EU (DWD), GFS (NOAA) and IFS HRES (ECMWF),
are fetched run by run from the Open-Meteo single-runs archive~\cite{openmeteo}. A run is admitted only if its
initialisation is at most the cut-off minus a lag, the provider's measured publication delay plus a margin (3.5\,h for
ICON-EU, 7\,h for GFS and IFS); at 11:30 this admits the 06 UTC ICON-EU run and the 00 UTC GFS and IFS runs, in summer and in winter time
(Figure~\ref{fig:nwp}). Since 19 October a run must also be
complete: if any variable is less than 99\% filled over lead times 12--48\,h, which contain the delivery day, the fetch
is retried up to five times and the builder then falls back to an earlier cycle, at most two
(Section~\ref{sec:limits}). Weather is sampled at 153 capacity-weighted $k$-means
centroids built from public registers (Table~\ref{tab:points}).
The table carries, per variable and bucket, each NWP model's capacity-weighted zone mean, and across the
three models the ensemble mean, spread and pairwise differences, so that disagreement can serve as a feature. Models
trained on the 28-day context train on tables built by the same rule day by day, from the runs that were available at
each day's own cut-off. The two-year training frames of \m{lgbm\_prod} and \m{lin\_cov\_2y} predate the single-runs
archive for ICON-EU and GFS; there the archive's previous-runs product is matched to the live lead time to the nearest
day, which gives forecasts of the same lead but not always of the same run (Section~\ref{sec:limits}). Figure~\ref{fig:map} shows the sampling points.

\begin{table}[t]
\centering\small
\caption{Weather sampling points per zone and bucket (capacity-weighted $k$-means centroids; load weighted by
population) and the public registers they are built from.}
\label{tab:points}
\begin{tabularx}{\linewidth}{@{}l c c c c Y@{}}
\toprule
Zone & \makecell{Onshore\\wind} & \makecell{Offshore\\wind} & Solar & Load & Registers \\
\midrule
BE & 8 & 2 & 8 & 6 & OpenStreetMap turbines scaled to Elia's capacity; RIVM offshore turbines; Fluvius and ORES municipal PV scaled to Elia; Statbel population \\
NL & 8 & 4 & 8 & 6 & RIVM turbines; CBS municipal PV and population \\
FR & 8 & 4 & 8 & 6 & ODRÉ national register of installations (commune level); INSEE population \\
DE-LU & 8 & 4 & 8 & 6 & Marktstammdatenregister; GeoNames population \\
DK1 & 8 & 4 & 8 & 6 & Energinet capacity per municipality; Danish Energy Agency offshore farms; GeoNames population \\
DK2 & 8 & 3 & 8 & 6 & as DK1 \\
\midrule
Total & 48 & 21 & 48 & 36 & 153 points \\
\bottomrule
\end{tabularx}
\end{table}

\paragraph{Load forecast.} ENTSO-E's day-ahead total load forecast (A65), which TSOs must publish two hours before the
gate. It is an input of the price models. By design the load models do not receive it, because in most zones it is itself
the pre-gate reference they are compared with (Section~\ref{sec:tso}); until issue day 15 October they did. Section~\ref{sec:limits}
records a parsing fault in its Danish series.

\paragraph{Fuel cost.} Daily closes of TTF gas (ICE Dutch front month) and of an exchange-traded EU carbon allowance
certificate, as published by Yahoo Finance, the last close dated two days before delivery or earlier, and the derived
short-run marginal cost of a combined-cycle gas plant in EUR/MWh,
\begin{equation}
\mathrm{SRMC} = \frac{P_{\mathrm{TTF}}}{\eta} + e\,P_{\mathrm{EUA}}, \qquad \eta = 0.52,\quad e = 0.37\,\mathrm{t\,CO_2/MWh},
\end{equation}
the fuel and carbon cost in the clean spark spread, with $\eta$ the plant's efficiency and $e$ its emission intensity
per MWh of electricity.

\subsection{Entries}

A model is a folder with a declarative \m{model.yaml}: a named sktime~\cite{sktime} estimator, a Python module, or a
subprocess script with its own environment, the route all foundation models take (each model loads its weights once
per run, for all series). A model whose environment is missing is skipped and logged, not failed. The most recent live
run, on 4 November 2026, took 17 minutes for the whole field and all 24 series on a four-core, 8\,GB server, within the
30-minute window; two models run in parallel. The field on 31 October (Table~\ref{tab:field}) has 43 active entries.

\begin{table}[!tp]
\centering\footnotesize
\setlength{\tabcolsep}{4pt}
\begin{threeparttable}
\caption{The field, 31 October 2026 (43 active entries). Parameter counts are those of the checkpoints the benchmark
loads. Covariates: \emph{native}, the model's own known-future covariate input; \emph{ridge}, an in-context ridge
regression on the covariates whose residual the model forecasts; \emph{full}, the whole covariate table.}
\label{tab:field}
\begin{tabularx}{\linewidth}{@{}l r l Y@{}}
\toprule
Entry & Params & Covariates & Developer, set-up \\
\midrule
\multicolumn{4}{@{}l}{\emph{Baselines}} \\
\m{naive\_weekly} & & & same quarter-hour one week earlier; the reference of all relative errors \\
\m{naive\_daily} & & & same quarter-hour one day earlier \\
\m{climatology} & & & same local time of day within $\pm14$ days, 2023 to August 2026 \\
\m{lin\_lags}, \m{lgbm\_lags} & & & ridge and LightGBM on the series' own lags; two years, retrained monthly \\
\addlinespace
\multicolumn{4}{@{}l}{\emph{Models trained on the covariates}\tnote{a}} \\
\m{lin\_cov}, \m{lgbm\_cov} & & full & ridge and LightGBM on the 28-day context \\
\m{lin\_cov\_2y} & & full & ridge on two years \\
\m{lgbm\_prod} & & full & LightGBM on two years, 120-day recency half-life, solar on its capacity factor\tnote{b} \\
\addlinespace
\multicolumn{4}{@{}l}{\emph{Foundation models, zero-shot}\tnote{c}} \\
Chronos-2~\cite{chronos} & 120M & native & Amazon \\
Chronos-Bolt base~\cite{chronos} & 205M & & Amazon \\
TimesFM 2.5~\cite{timesfm} & 231M & ridge & Google \\
TimesFM 3.0~\cite{timesfm} & 331M & native & Google \\
Moirai 2 small~\cite{moirai} & 11M & \tnote{d} & Salesforce \\
TiRex~\cite{tirex} & 35M & & NX-AI \\
TiRex-2~\cite{tirex} & 95M & native & NX-AI; about 38M active in univariate mode \\
Toto & 151M & \tnote{d} & Datadog; retired 27 October, superseded by Toto 2 \\
Toto 2 & 313M, 1B & & Datadog; live from 25 October \\
Sundial & 128M & & Tsinghua University \\
TTM r2 & 3.4M & & IBM \\
Tabby, Tabby-Pretrain & 147M, 146M & ridge & Huawei Noah's Ark Lab; load forecast without covariates \\
Timestral & 209M & native & Mistral AI \\
TimeH2O & 108M & native & H2O.ai; live from 28 October, with covariates from 29 October \\
t0-alpha, t0-beta, t0-gamma & 102M, 256M, 256M & native & The Forecasting Company \\
\addlinespace
\multicolumn{4}{@{}l}{\emph{Ensembles: quarter-hourly median of fixed members, point and each quantile; live from 20 October}} \\
\m{ens\_fm} & & & the newest history-only model of five native-covariate families \\
\m{ens\_fm\_cov} & & native & their five covariate variants \\
\m{ens\_hybrid} & & native & those five and \m{lgbm\_prod} \\
\addlinespace
\multicolumn{4}{@{}l}{\emph{References: the grid operators' forecasts}} \\
\m{tso\_cutoff} & & & as published before the gate; rated only where it covers every selected series; live only\tnote{e} \\
\m{tso\_final} & & & final day-ahead wind and solar, published after the gate; never ranked\tnote{f} \\
\bottomrule
\end{tabularx}
\begin{tablenotes}\footnotesize
\item[a] \emph{Full}: every NWP model's columns, the ensemble statistics and differences, the calendar and, for price,
yesterday's price. The foundation models take a compact subset instead: three or four ensemble means and spreads per
target, and for price the load forecast, fuel cost and the renewable drivers of the coupled zones.
\item[b] Retrained daily for price and load, monthly for wind and solar.
\item[c] Every foundation model runs on the series' own history; those with a covariate entry run a second time with
the frozen covariates (e.g.\ \m{chronos2} and \m{chronos2\_cov}).
\item[d] Covariate variants retired after trial runs: covariates made Moirai~2 worse, and Toto gained too little for
its run time and memory.
\item[e] Elia's 11:00 day-ahead wind and solar forecast for BE, or, where that field is still empty at the run, Elia's most recent forecast read at the run; NED's national wind, solar and load forecast for NL,
fetched at the cut-off; the ENTSO-E day-ahead load forecast elsewhere.
\item[f] Elia 18:00, RTE 16:15, ENTSO-E for DE-LU and Danish wind, NED in the evening for NL.
\end{tablenotes}
\end{threeparttable}
\end{table}

No TSO in France or Germany publishes a day-ahead wind or solar forecast before the gate (RTE at 16:15, the German
TSOs at 18:00 on $D-1$), and Energinet's is not verified as published before it, so \m{tso\_cutoff} has wind and solar only for BE and NL. Classical
univariate models (ARIMA, ETS, Theta, CES, TBATS) were in the field in its first two weeks and were retired: on
15-minute data with a 96-step daily cycle and a 28-day context they were not competitive. Their records are kept but
no longer shown.

\subsection{Scoring}

For every forecast, issue day $d$ and series $s$ the benchmark records MAE, RMSE and bias and, for models with deciles,
the Winkler score and coverage of the 80\% interval and the weighted interval score (WIS). For the median and the four
central intervals that the nine deciles define, the WIS equals twice the pinball loss averaged over the
deciles~\cite{bracher},
\begin{equation}
\mathrm{WIS} = \frac{2}{9\,T}\sum_{t=1}^{T}\sum_{\tau\in\{0.1,\dots,0.9\}} \max\!\big(\tau\,(y_t - q_{\tau,t}),\,(\tau-1)(y_t - q_{\tau,t})\big),
\end{equation}
with $T$ the scored quarter-hours of the day. Errors are compared across series of different scale by relating them
to the baseline \m{naive\_weekly}, and they are aggregated \emph{before} they are normalised: for model $m$, series
$s$ and the set $\mathcal{D}_{m,s}$ of issue days on which both $m$ and the baseline were scored,
\begin{equation}
\mathrm{RelMAE}_{m,s} = \frac{\sum_{d\in\mathcal{D}_{m,s}} \sum_t |e^{m}_{d,t}|}{\sum_{d\in\mathcal{D}_{m,s}} \sum_t |e^{\text{base}}_{d,t}|},
\qquad
\mathrm{RelRMSE}_{m,s} = \Big(\frac{\sum_{d,t} (e^{m}_{d,t})^2}{\sum_{d,t} (e^{\text{base}}_{d,t})^2}\Big)^{1/2},
\end{equation}
and Rel.\ WIS likewise as the ratio of the summed daily WIS of model and baseline.
A ratio of means rather than a mean of daily ratios keeps one calm day, on which the baseline happened to be nearly
right, from dominating the average. Over several series the per-series values are combined by the mean, the median or
the geometric mean, selectable on the benchmark's website. The geometric mean, used in the tables below, treats halving and
doubling the error symmetrically and penalises a model that fails on one series.

Three choices in these definitions affect which model is ranked best, and they are stated here explicitly.
\emph{The functional rewarded.} Absolute error is minimised by the median of the
predictive distribution and squared error by its mean~\cite{gneiting2011}, and for skewed targets such as solar power
or spiking prices the two differ. Most foundation models report the median of their predictive quantiles (Chronos,
TiRex, Moirai, Sundial, Toto, t0, Timestral, Tabby). TimesFM, the climatology and the LightGBM and ridge entries,
fitted on squared error, report a mean. The standings use MAE and therefore favour the median. Rel.\ RMSE, which
favours the mean, is reported alongside. Bias is reported separately, since neither measure separates a systematic
offset from random error. \emph{The benchmark in the denominator.} Every scale-free error measure is defined relative to a
benchmark forecast~\cite{hyndmankoehler}. EnergyBench uses the value of the same quarter-hour one week earlier, a
seasonal naive forecast that reproduces the daily and weekly cycles of price and load and requires no fitting. Unlike the
scaling of the MASE, which uses the naive forecast's error on the training data, the baseline's error is taken on the
same days as the model's, so that both face the same conditions~\cite{davydenko}. For wind and solar this benchmark is weak, and relative errors well below one
are to be expected. It serves as a common unit of comparison and is not a measure of how difficult a series is to
forecast. \emph{Aggregation.} Errors are averaged over the 96 quarter-hours of the delivery day and over issue days
within a series, and then across series with equal weight per series. With equal weights, the standings measure
performance across forecasting problems rather than where most energy is at stake.

\subsection{Ranking}
\label{sec:ranking}

A \emph{race} is one issue day and one series; its field is every model with a scored forecast in it. A model whose
run failed is left out of that race rather than ranked last (the run tracker on the site shows every failure), and a
model joins the races of the series it declares from the day it is added. Leaving failures out is defensible only
because the benchmark runs every entry itself: an entry cannot choose the days on which it forecasts. A benchmark that
accepted submitted forecasts would have to count a missing forecast as a loss or give it a penalty score, as Predico does by assigning it the 75th
percentile of that day's scores~\cite{predico}, since a participant could otherwise withhold forecasts on difficult days. The grid operators' pre-gate forecast, which
covers only some series, is rated only on selections whose every series it forecasts (for example BE wind, or load); on
any other selection it takes no part in the races, so that no rating rests on a different set of series from the
rest of the field, and its errors are shown without a rating. Within a race every pair of models plays
one game, won by the lower relative error of the day (equal errors split the game). All games of the selected period
are fitted at once with a Bradley--Terry model~\cite{bradleyterry}, $P(i \text{ beats } j) = p_i/(p_i+p_j)$, by the
minorisation--maximisation algorithm of Hunter~\cite{hunter}, and reported on the Elo scale,
$\mathrm{Elo}_i = 1000 + 400\log_{10} p_i$, with the geometric mean of the strengths fixed at 1 so that 1000 is the
field's average. One virtual drawn game against an average opponent keeps the rating finite for a model that wins or
loses every game. The construction follows rating systems for chess and sports~\cite{elo1978}. Unlike a running Elo, the rating does
not depend on the order of the races, and a win over a strong
model counts more than a win over a weak one; the same construction underlies the Chatbot Arena
leaderboard~\cite{chatbotarena}.

The 95\% interval is the 2.5th--97.5th percentile of 200 refits on issue days drawn with replacement. Days rather than
races are resampled because all races of a day share the same weather and tend to be won and lost together.
Overlapping intervals mean the order of two models is not settled. Any pairwise rating is relative to the field: a model's score depends on whom it
played, and several near-identical entries (for example three releases of one model family) take wins from each other and
from everyone else. The pooled relative errors do not depend on the field and are always reported next to the rating. A model is listed as provisional if it has fewer races than the smaller of 12 and a fifth of the most
any model has. The rating can be computed for any selection of zones, targets and issue days. The rating counts
wins and not their margins (Section~\ref{sec:limits}).

\subsection{Operation and reproducibility}

The benchmark runs as scheduled jobs on one server: the live run at 11:30, scoring at
07:00, 14:45 and 20:45, an hourly fetch of provisional actuals, and a catch-up at 03:30. Every forecast, input
snapshot, covariate table, provenance file and revision log is kept, and the website is a static export of them, so
any number on it can be recomputed. The configuration file that defines the benchmark (cut-off, gate, resolution,
deciles, context, baseline, sources, admission lags) is versioned with the code: changing it changes
the benchmark. The website and this paper report live results only.

\section{Results: the first thirty-two live days}
\label{sec:results}

The results below are those of the live record only, scored on the forecasts as they were locked before each gate.
After the coverage of the record, they give the standings, the value of the covariates for foundation models, the
role of model size, and the comparison with purpose-built models and with the forecasts issued by the grid operators.

\subsection{Coverage}
\label{sec:coverage}

\begin{figure}[t]
\centering
\begin{tikzpicture}[x=0.24cm,y=0.42cm,font=\scriptsize]
\draw[gray!25] (2.5,-18.50) -- (2.5,1.4);
\draw[gray!25] (5.5,-18.50) -- (5.5,1.4);
\draw[gray!25] (10.5,-18.50) -- (10.5,1.4);
\draw[gray!25] (15.5,-18.50) -- (15.5,1.4);
\draw[gray!25] (20.5,-18.50) -- (20.5,1.4);
\draw[gray!25] (25.5,-18.50) -- (25.5,1.4);
\draw[gray!25] (30.5,-18.50) -- (30.5,1.4);
\draw[gray!25] (33.5,-18.50) -- (33.5,1.4);
\node[anchor=west,font=\scriptsize\bfseries] at (2,0.75) {Series};
\node[anchor=east] at (1.7,0.00) {BE, FR: price, wind, solar};
\path[fill=histc!70] (2,-0.30) rectangle (34,0.30);
\node[anchor=west] at (34.2,0.00) {32};
\node[anchor=east] at (1.7,-1.00) {NL: price};
\path[fill=histc!70] (2,-1.30) rectangle (34,-0.70);
\node[anchor=west] at (34.2,-1.00) {32};
\node[anchor=east] at (1.7,-2.00) {BE, FR: load};
\path[fill=histc!70] (6,-2.30) rectangle (34,-1.70);
\node[anchor=west] at (34.2,-2.00) {28};
\node[anchor=east] at (1.7,-3.00) {NL: wind, solar; load};
\path[fill=histc!70] (11,-3.30) rectangle (34,-2.70);
\node[anchor=west] at (34.2,-3.00) {23; 22};
\node[anchor=east] at (1.7,-4.00) {DE-LU; DK1, DK2: price, solar};
\path[fill=histc!70] (15,-4.30) rectangle (34,-3.70);
\node[anchor=west] at (34.2,-4.00) {19};
\node[anchor=east] at (1.7,-5.00) {DK1, DK2: wind};
\path[fill=histc!70] (15,-5.30) rectangle (17,-4.70);
\path[fill=histc!70] (19,-5.30) rectangle (24,-4.70);
\path[fill=histc!70] (25,-5.30) rectangle (34,-4.70);
\node[anchor=west] at (34.2,-5.00) {16};
\node[anchor=east] at (1.7,-6.00) {DK1, DK2: load};
\path[pattern=north east lines, pattern color=histc] (15,-6.30) rectangle (34,-5.70);
\node[anchor=west] at (34.2,-6.00) {0};
\node[anchor=west,font=\scriptsize\bfseries] at (2,-7.05) {Entries};
\node[anchor=east,align=right] at (1.7,-7.90) {\m{naive\_daily}, \m{naive\_weekly}, Chronos-2, \m{lgbm\_lags}};
\path[fill=covc!75] (2,-8.20) rectangle (34,-7.60);
\node[anchor=west] at (34.2,-7.90) {32};
\node[anchor=east,align=right] at (1.7,-9.15) {\begin{tabular}[c]{@{}r@{}}Chronos-2 cov., Chronos-Bolt, Moirai 2, Sundial, TimesFM 2.5\\ and 3.0 (both $\pm$cov.), TiRex, TTM r2, \m{lgbm\_cov}, climatology\end{tabular}};
\path[fill=covc!75] (3,-9.45) rectangle (34,-8.85);
\node[anchor=west] at (34.2,-9.15) {31};
\node[anchor=east,align=right] at (1.7,-10.40) {Toto};
\path[fill=covc!75] (3,-10.70) rectangle (28,-10.10);
\node[anchor=west] at (34.2,-10.40) {25};
\node[anchor=east,align=right] at (1.7,-11.40) {t0-alpha, t0-beta ($\pm$cov.); \m{tso\_cutoff}};
\path[fill=covc!75] (5,-11.70) rectangle (34,-11.10);
\node[anchor=west] at (34.2,-11.40) {29};
\node[anchor=east,align=right] at (1.7,-12.40) {TiRex-2 ($\pm$cov.)};
\path[fill=covc!75] (8,-12.70) rectangle (34,-12.10);
\node[anchor=west] at (34.2,-12.40) {26};
\node[anchor=east,align=right] at (1.7,-13.40) {\m{lin\_lags}, \m{lin\_cov}, \m{lin\_cov\_2y}};
\path[fill=covc!75] (14,-13.70) rectangle (34,-13.10);
\node[anchor=west] at (34.2,-13.40) {20};
\node[anchor=east,align=right] at (1.7,-14.65) {\begin{tabular}[c]{@{}r@{}}t0-gamma, Timestral, Tabby, Tabby-Pretrain ($\pm$cov.),\\ \m{lgbm\_prod}\end{tabular}};
\path[fill=covc!75] (16,-14.95) rectangle (34,-14.35);
\node[anchor=west] at (34.2,-14.65) {18};
\node[anchor=east,align=right] at (1.7,-15.90) {\m{ens\_fm}, \m{ens\_fm\_cov}, \m{ens\_hybrid}};
\path[fill=covc!75] (20,-16.20) rectangle (34,-15.60);
\node[anchor=west] at (34.2,-15.90) {14};
\node[anchor=east,align=right] at (1.7,-16.90) {Toto 2, Toto 2 1B};
\path[fill=covc!75] (25,-17.20) rectangle (34,-16.60);
\node[anchor=west] at (34.2,-16.90) {9};
\node[anchor=east,align=right] at (1.7,-17.90) {TimeH2O ($\pm$cov.), not ranked};
\path[pattern=north east lines, pattern color=covc] (28,-18.20) rectangle (34,-17.60);
\node[anchor=west] at (34.2,-17.90) {6};
\node[anchor=north,gray] at (2.5,-18.50) {2};
\node[anchor=north,gray] at (5.5,-18.50) {5};
\node[anchor=north,gray] at (10.5,-18.50) {10};
\node[anchor=north,gray] at (15.5,-18.50) {15};
\node[anchor=north,gray] at (20.5,-18.50) {20};
\node[anchor=north,gray] at (25.5,-18.50) {25};
\node[anchor=north,gray] at (30.5,-18.50) {30};
\node[anchor=north,gray] at (33.5,-18.50) {2 Nov};
\node[anchor=north] at (18,-19.40) {issue day, October--November 2026};
\node[anchor=west,gray] at (34.2,1.15) {days};
\end{tikzpicture}
\caption{The live record, issue days 2 October--2 November 2026. Top: when each series is scored from, with its
number of scored issue days (Danish wind lost three; Danish load is forecast but left out, Section~\ref{sec:coverage}).
Bottom: when each entry first forecast, with the issue days it ran; Toto left after 27 October. TimeH2O, with six days,
is not yet ranked.}
\label{fig:record}
\end{figure}
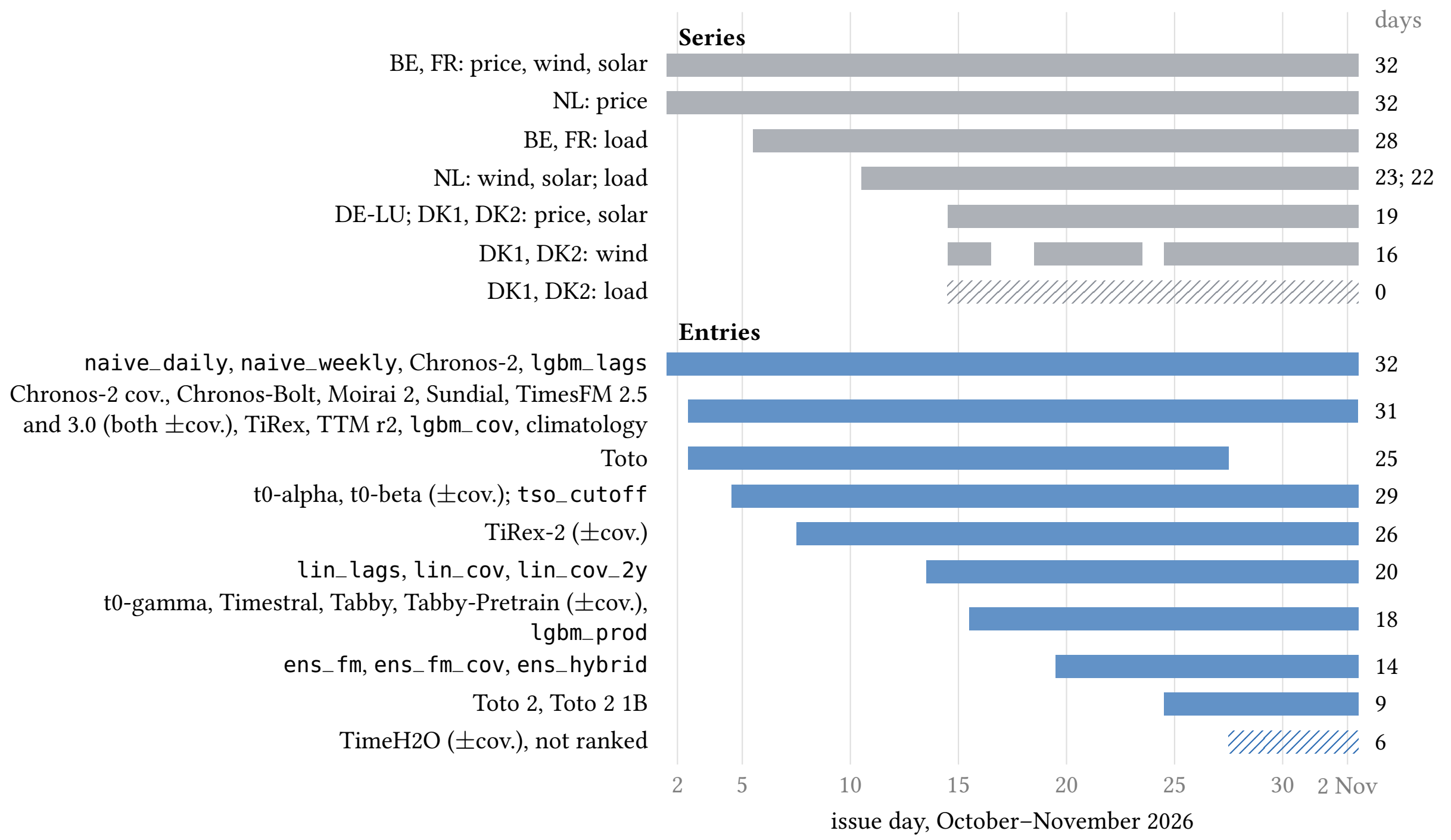

The live record covers issue days 2~October--2~November 2026, the last day with every target scored when this paper was
written. Series enter the record when they can be scored on a stable definition, and entries when they are added, so
the record is uneven (Figure~\ref{fig:record}). The Dutch wind, solar and load record starts on 11--12 October,
with NED's national series (Section~\ref{sec:actuals}), and DE-LU, DK1 and DK2 join on 15 October.\footnote{Danish
wind lost three issue days: 17 and 18 October to its own change of source, and 24 October, whose delivery day was that
of the change to winter time, to a missing actual.} ENTSO-E has published Danish load only in part since 10 October,
which leaves twelve scored days, none since 28 October, and the pre-gate reference covered only part of the delivery day on ten zone-days
(Section~\ref{sec:limits}). A series with so few days would weigh as much as any other in the averages below, so
Danish load is left out. The record therefore holds 22 of the 24 series, with between 16 and 32 issue days each.
Because the entries also joined at different times, every comparison below is made on the races the compared models
share.

\subsection{Standings}
\label{sec:standings}

\begin{figure}[t]
\centering
\begin{tikzpicture}
\begin{axis}[
  width=0.86\linewidth, height=15.9cm,
  xmin=680, xmax=1340, ymin=-0.7, ymax=39.7,
  ytick={0,...,39}, yticklabels from table={elo_live.dat}{label},
  yticklabel style={font=\ttfamily\scriptsize},
  xlabel={Elo (live, six zones, 2 October--2 November 2026)}, xlabel style={font=\small},
  tick label style={font=\scriptsize}, xmajorgrids, grid style={gray!20},
  axis line style={gray!60}, tick style={gray!60},
  extra x ticks={1000}, extra x tick style={grid style={black!50,dashed}, tick label style={opacity=0}},
]
\addplot[only marks, mark=*, mark size=1.7pt, histc, error bars/.cd, x dir=both, x explicit, error bar style={histc}]
  table[x=elo, y=y, x error plus=hi, x error minus=lo, restrict expr to domain={\thisrow{cov}}{0:0}] {elo_live.dat};
\addplot[only marks, mark=*, mark size=1.9pt, covc, error bars/.cd, x dir=both, x explicit, error bar style={covc, thick}]
  table[x=elo, y=y, x error plus=hi, x error minus=lo, restrict expr to domain={\thisrow{cov}}{1:1}] {elo_live.dat};
\end{axis}
\end{tikzpicture}
\caption{Live Elo ratings with day-bootstrap 95\% intervals, entries of Table~\ref{tab:live}. Blue: models using the
frozen covariates; grey: history only. The pre-gate TSO forecast is not rated over all series
(Section~\ref{sec:results}). The dashed line is the field average.}
\label{fig:elo}
\end{figure}
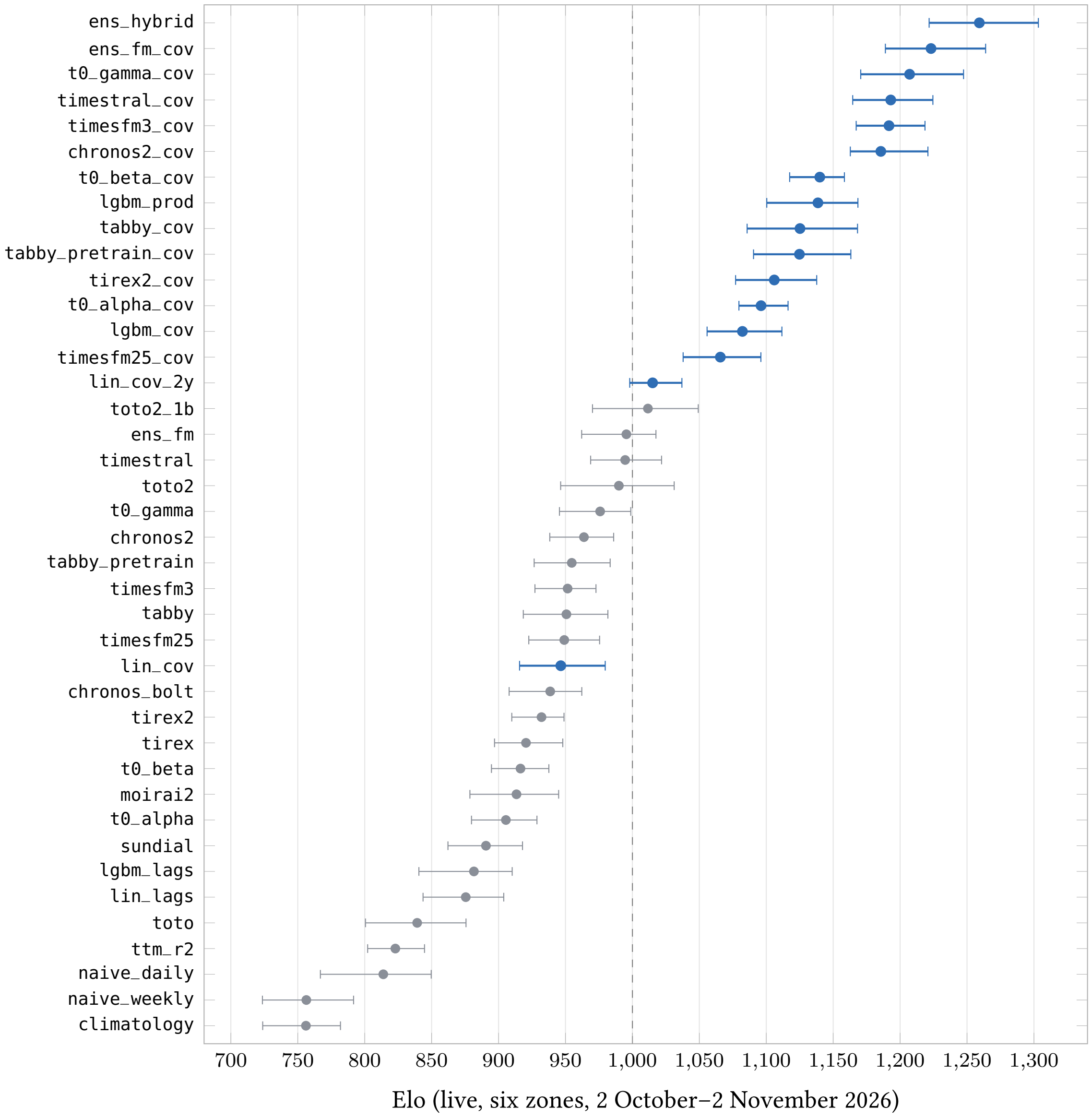

Table~\ref{tab:live} gives the standings for the 40 ranked entries, and Figure~\ref{fig:elo} shows their ratings.
TimeH2O, live since 28 October, is left out of the ranking for now: over its five days with covariates its rating
interval is about twice as wide as those of the leading models. The two median ensembles of covariate models, live
since 20 October, rate highest (\m{ens\_hybrid}, Elo 1259 and Rel.\ MAE 0.240; \m{ens\_fm\_cov}, 1223 and 0.247). Among
single models, t0-gamma, Timestral, TimesFM 3.0 and Chronos-2 with covariates lead (Elo 1186--1207, Rel.\ MAE
0.273--0.295), with overlapping intervals. Behind the leading group the field splits in two. Every covariate-aware
foundation model except TimesFM 2.5 is separated by its interval from every history-only model, and the best
history-only entries, Toto 2 at 1B parameters over its nine days and the history-only ensemble (Elo 1012 and 996),
rank behind fifteen covariate models. Among the history-only foundation models the differences are small (pooled
Rel.\ MAE 0.51--0.79, leaving out Toto 2) compared with the effect of giving any of them the weather (0.27--0.36 with
covariates, Table~\ref{tab:cov}).

The ensembles are medians with members fixed in advance by a rule, not by score (Table~\ref{tab:field}), so they do
not require knowing which member will perform best. On the fourteen issue days on which they ran (20 October--2
November, 306 races each), \m{ens\_fm\_cov} has a lower error than each of its five members (0.247, against
0.251--0.288), and \m{ens\_hybrid}, which adds \m{lgbm\_prod}, a model of a different class trained on the same
weather, has a lower error than every single model (0.240, against 0.251 for the best, TimesFM 3.0). Fourteen days
are, however, too few to rank the ensembles reliably against their best members.

The grid operators' pre-gate forecast (\m{tso\_cutoff}) has no rating in these standings. It covers only 8 of the
22 series, so a rating would rest on games in those series alone, and under the rule of Section~\ref{sec:ranking} it
takes no part in the races of a selection it does not fully cover; the ratings of the other entries are fitted without
it. It is listed below the ranked entries with its pooled errors, which are a mean over a different set of series,
weighted towards load, and cannot be compared directly with those of the entries that forecast all 22. A rating would
also count wins rather than margins: on wind and solar, \m{lgbm\_prod} has the lower error in only 44\% of the races
it shares with the operators, while its pooled error there is level with theirs. Section~\ref{sec:tso} therefore
compares the operators with each entry on the races both forecast.

\begin{table}[!t]
\centering\scriptsize
\setlength{\tabcolsep}{4pt}
\begin{threeparttable}
\caption{Live standings, issue days 2 October--2 November 2026, all six zones (22 series). Elo with its day-bootstrap 95\%
interval; Rel.\ MAE and Rel.\ RMSE pooled per series against \m{naive\_weekly}, geometric mean over the series a model
forecasts (``all'') or over the zones of one target. Ranked entries; covariate models in bold.}
\label{tab:live}
\begin{tabular}{@{}rlrcrrrrrrr@{}}
\toprule
& Model & Elo & 95\% interval & Races & Rel.\ MAE & Rel.\ RMSE & Price & Wind & Solar & Load \\
\cmidrule(lr){6-7}\cmidrule(l){8-11}
& & & & & \multicolumn{2}{c}{all series} & \multicolumn{4}{c}{Rel.\ MAE per target} \\
\midrule
% rows generated by scripts/research/preprint_all.py (as-run scores, issue days up to 2026-11-02)
    1 & \texttt{\textbf{ens\_hybrid}} & 1259 & [1222, 1303] & 306 & 0.240 & 0.276 & 0.195 & 0.130 & 0.314 & 0.546 \\
    2 & \texttt{\textbf{ens\_fm\_cov}} & 1223 & [1189, 1264] & 306 & 0.247 & 0.285 & 0.197 & 0.136 & 0.324 & 0.569 \\
    3 & \texttt{\textbf{t0\_gamma\_cov}} & 1207 & [1171, 1247] & 390 & 0.275 & 0.311 & 0.224 & 0.162 & 0.344 & 0.586 \\
    4 & \texttt{\textbf{timestral\_cov}} & 1193 & [1165, 1225] & 390 & 0.273 & 0.306 & 0.250 & 0.151 & 0.321 & 0.597 \\
    5 & \texttt{\textbf{timesfm3\_cov}} & 1192 & [1167, 1219] & 525 & 0.282 & 0.307 & 0.229 & 0.160 & 0.336 & 0.686 \\
    6 & \texttt{\textbf{chronos2\_cov}} & 1186 & [1163, 1221] & 525 & 0.295 & 0.329 & 0.228 & 0.184 & 0.367 & 0.637 \\
    7 & \texttt{\textbf{t0\_beta\_cov}} & 1140 & [1117, 1158] & 511 & 0.320 & 0.356 & 0.263 & 0.196 & 0.392 & 0.657 \\
    8 & \texttt{\textbf{lgbm\_prod}} & 1139 & [1100, 1169] & 390 & 0.293 & 0.314 & 0.262 & 0.143 & 0.381 & 0.685 \\
    9 & \texttt{\textbf{tabby\_cov}} & 1125 & [1086, 1168] & 390 & 0.313 & 0.340 & 0.259 & 0.192 & 0.385 & 0.634 \\
    10 & \texttt{\textbf{tabby\_pretrain\_cov}} & 1125 & [1091, 1163] & 390 & 0.314 & 0.342 & 0.260 & 0.191 & 0.375 & 0.668 \\
    11 & \texttt{\textbf{tirex2\_cov}} & 1106 & [1077, 1138] & 486 & 0.315 & 0.336 & 0.271 & 0.164 & 0.398 & 0.745 \\
    12 & \texttt{\textbf{t0\_alpha\_cov}} & 1096 & [1080, 1116] & 511 & 0.334 & 0.359 & 0.326 & 0.171 & 0.390 & 0.749 \\
    13 & \texttt{\textbf{lgbm\_cov}} & 1082 & [1056, 1112] & 525 & 0.337 & 0.355 & 0.300 & 0.154 & 0.396 & 1.026 \\
    14 & \texttt{\textbf{timesfm25\_cov}} & 1066 & [1038, 1096] & 525 & 0.362 & 0.389 & 0.302 & 0.194 & 0.418 & 0.978 \\
    15 & \texttt{\textbf{lin\_cov\_2y}} & 1015 & [998, 1037] & 424 & 0.389 & 0.403 & 0.273 & 0.184 & 0.474 & 1.514 \\
    16 & \texttt{toto2\_1b} & 1012 & [970, 1049] & 198 & 0.465 & 0.523 & 0.245 & 0.443 & 0.743 & 0.647 \\
    17 & \texttt{ens\_fm} & 996 & [962, 1018] & 306 & 0.486 & 0.526 & 0.302 & 0.495 & 0.665 & 0.606 \\
    18 & \texttt{timestral} & 995 & [969, 1022] & 390 & 0.511 & 0.545 & 0.318 & 0.541 & 0.686 & 0.613 \\
    19 & \texttt{toto2} & 990 & [946, 1031] & 198 & 0.477 & 0.535 & 0.256 & 0.429 & 0.762 & 0.703 \\
    20 & \texttt{t0\_gamma} & 976 & [946, 999] & 390 & 0.527 & 0.561 & 0.323 & 0.531 & 0.726 & 0.675 \\
    21 & \texttt{chronos2} & 964 & [938, 986] & 532 & 0.571 & 0.609 & 0.350 & 0.563 & 0.850 & 0.667 \\
    22 & \texttt{tabby\_pretrain} & 955 & [927, 983] & 390 & 0.546 & 0.581 & 0.346 & 0.574 & 0.714 & 0.668 \\
    23 & \texttt{timesfm3} & 952 & [927, 973] & 525 & 0.578 & 0.608 & 0.344 & 0.584 & 0.855 & 0.690 \\
    24 & \texttt{tabby} & 951 & [918, 982] & 390 & 0.552 & 0.588 & 0.339 & 0.575 & 0.789 & 0.634 \\
    25 & \texttt{timesfm25} & 949 & [923, 975] & 525 & 0.573 & 0.609 & 0.357 & 0.551 & 0.799 & 0.746 \\
    26 & \texttt{\textbf{lin\_cov}} & 946 & [916, 980] & 424 & 0.451 & 0.469 & 0.367 & 0.205 & 0.580 & 1.372 \\
    27 & \texttt{chronos\_bolt} & 939 & [908, 962] & 525 & 0.587 & 0.623 & 0.358 & 0.562 & 0.887 & 0.710 \\
    28 & \texttt{tirex2} & 932 & [910, 949] & 486 & 0.572 & 0.599 & 0.359 & 0.536 & 0.841 & 0.716 \\
    29 & \texttt{tirex} & 921 & [897, 948] & 525 & 0.600 & 0.630 & 0.375 & 0.571 & 0.801 & 0.850 \\
    30 & \texttt{t0\_beta} & 916 & [895, 938] & 511 & 0.588 & 0.615 & 0.390 & 0.544 & 0.871 & 0.675 \\
    31 & \texttt{moirai2} & 913 & [879, 945] & 525 & 0.604 & 0.643 & 0.382 & 0.546 & 0.889 & 0.786 \\
    32 & \texttt{t0\_alpha} & 906 & [880, 929] & 511 & 0.596 & 0.624 & 0.382 & 0.559 & 0.840 & 0.767 \\
    33 & \texttt{sundial} & 891 & [862, 918] & 525 & 0.622 & 0.635 & 0.400 & 0.579 & 0.924 & 0.745 \\
    34 & \texttt{lgbm\_lags} & 882 & [840, 910] & 532 & 0.662 & 0.688 & 0.473 & 0.562 & 0.899 & 0.886 \\
    35 & \texttt{lin\_lags} & 876 & [844, 904] & 424 & 0.612 & 0.608 & 0.417 & 0.495 & 0.906 & 0.830 \\
    36 & \texttt{toto} & 839 & [801, 876] & 393 & 0.790 & 0.813 & 0.556 & 0.763 & 0.966 & 1.040 \\
    37 & \texttt{ttm\_r2} & 823 & [802, 845] & 525 & 0.677 & 0.663 & 0.428 & 0.526 & 0.913 & 1.255 \\
    38 & \texttt{naive\_daily} & 814 & [767, 850] & 532 & 0.800 & 0.863 & 0.476 & 0.632 & 0.955 & 1.908 \\
    39 & \texttt{naive\_weekly} & 756 & [724, 792] & 532 & 1.000 & 1.000 & 1.000 & 1.000 & 1.000 & 1.000 \\
    40 & \texttt{climatology} & 756 & [724, 782] & 525 & 0.988 & 0.949 & 1.170 & 0.628 & 0.867 & 1.844 \\
    \midrule
    \multicolumn{11}{@{}l}{\emph{References: the grid operators' forecasts (Section~\ref{sec:tso})}} \\
    -- & \texttt{tso\_cutoff} & -- & -- & 198 & 0.378 & 0.395 & -- & 0.148 & 0.378 & 0.605 \\
    -- & \texttt{tso\_final} & -- & -- & -- & 0.184 & 0.202 & -- & 0.118 & 0.285 & -- \\
\bottomrule
\end{tabular}
\begin{tablenotes}\footnotesize
\item Races differ because entries joined at different times (Toto 2, 25 October; ensembles 20 October; t0-gamma,
Timestral, Tabby, \m{lgbm\_prod} 16 October) or left (Toto, 27 October). The references are not ranked. \m{tso\_cutoff} forecasts 8 of the 22 series (BE and NL wind and solar; load
in BE, NL, FR and DE-LU), so it has no Elo over all series, and its pooled errors are not comparable with those of the
ranked entries (Section~\ref{sec:results}). \m{tso\_final}, published after the gate, is shown for wind and solar in BE, NL, FR and
DE-LU; its Danish wind forecasts cover a different scope from the actuals and are left out (Section~\ref{sec:tso}).
TimeH2O, live from 28 October, is not yet ranked (Section~\ref{sec:standings}).
\end{tablenotes}
\end{threeparttable}
\end{table}

\begin{figure}[t]
\centering
\includegraphics[width=\linewidth]{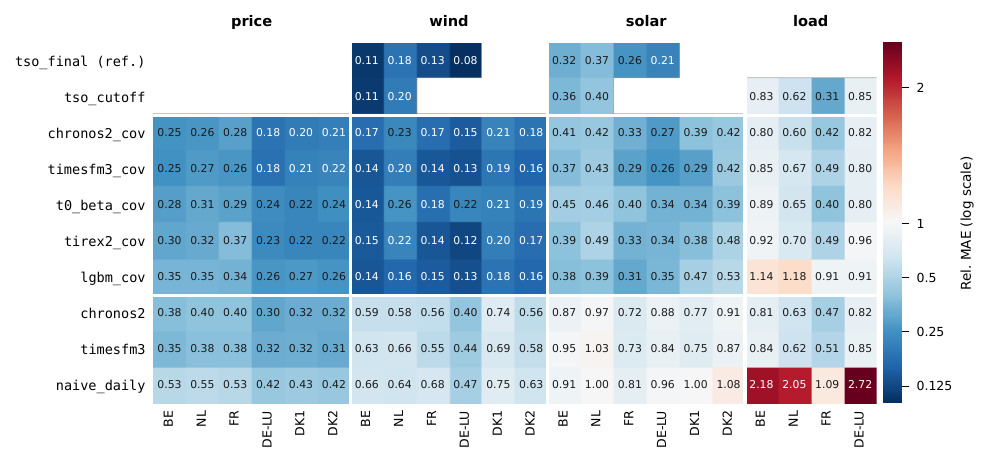}
\caption{Pooled Rel.\ MAE per series for the references, five covariate models, two history-only foundation models
and \m{naive\_daily} (live, 2 October--2 November 2026; colour on a log scale, 1 is the seasonal-naive baseline). Empty cells:
no forecast, or (for \m{tso\_final} in DK1 and DK2) a forecast of a different scope.}
\label{fig:heatmap}
\end{figure}

Figure~\ref{fig:heatmap} breaks the pooled error down by series. The covariate models lead on wind and solar in every
zone, by a margin much larger than the differences between them; Dutch and eastern Danish (DK2) solar are the hardest of these
series. On load the
history-only foundation models are level with the covariate ones, and the grid operators' pre-gate load forecast is
strong only in France.

\subsection{Effect of covariates on foundation models}

Ten foundation models run both on the series' own history and with the frozen covariate table, so the value of the
weather can be measured model by model on exactly the same races (Table~\ref{tab:cov}; TimeH2O is too recent to include). All ten gain from the covariates. For the seven that take covariates natively, the pooled relative error
falls by 44--51\%. The reduction is 64--73\% on wind and 53--61\% on solar, where a model that sees only the last four
weeks of output has no information about tomorrow's wind or cloud cover, and 15--34\% on price, through the load
forecast, fuel cost and the expected renewable infeed. On load the change is small ($-13$\% to $+4$\%), since a load
curve is dominated by daily and weekly cycles that the history already contains. Of the three models that receive the
covariates through an in-context ridge regression, the two Tabby models gain nearly as much (43\%). TimesFM 2.5 gains less (37\%), and its load error rises by 31\%, because the ridge regression fitted on
28 days extrapolates poorly.

\begin{table}[t]
\centering\small
\setlength{\tabcolsep}{4.5pt}
\caption{The same foundation model without and with the frozen covariates, on the races both forecast (live, all
zones, issue days 2 October--2 November 2026). ``Native'': the model takes known-future covariates itself; ``ridge'': an
in-context ridge regression on the covariates, the model forecasting its residual (Tabby forecasts load without
covariates). Rel.\ MAE pooled per series, geometric mean over series; a change is the geometric mean of the
per-series ratios minus one.}
\label{tab:cov}
\begin{tabular}{@{}llrrrrrrrrr@{}}
\toprule
& & & & \multicolumn{2}{c}{Rel.\ MAE} & \multicolumn{5}{c}{Change with covariates} \\
\cmidrule(lr){5-6}\cmidrule(l){7-11}
Model & Covariates & Days & Races & without & with & all & price & wind & solar & load \\
\midrule
Chronos-2 & native & 31 & 525 & 0.569 & 0.295 & $-48\%$ & $-34\%$ & $-67\%$ & $-57\%$ & $-4\%$ \\
TimesFM 3.0 & native & 31 & 525 & 0.578 & 0.282 & $-51\%$ & $-33\%$ & $-73\%$ & $-61\%$ & $-1\%$ \\
t0-alpha & native & 29 & 511 & 0.596 & 0.334 & $-44\%$ & $-15\%$ & $-69\%$ & $-54\%$ & $-2\%$ \\
t0-beta & native & 29 & 511 & 0.588 & 0.320 & $-46\%$ & $-33\%$ & $-64\%$ & $-55\%$ & $-3\%$ \\
t0-gamma & native & 18 & 390 & 0.527 & 0.275 & $-48\%$ & $-31\%$ & $-69\%$ & $-53\%$ & $-13\%$ \\
TiRex-2 & native & 26 & 486 & 0.572 & 0.315 & $-45\%$ & $-24\%$ & $-69\%$ & $-53\%$ & $+4\%$ \\
Timestral & native & 18 & 390 & 0.511 & 0.273 & $-46\%$ & $-21\%$ & $-72\%$ & $-53\%$ & $-3\%$ \\
\midrule
TimesFM 2.5 & ridge & 31 & 525 & 0.573 & 0.362 & $-37\%$ & $-15\%$ & $-65\%$ & $-48\%$ & $+31\%$ \\
Tabby & ridge & 18 & 390 & 0.552 & 0.313 & $-43\%$ & $-24\%$ & $-67\%$ & $-51\%$ & $+0\%$ \\
Tabby-Pretrain & ridge & 18 & 390 & 0.546 & 0.314 & $-43\%$ & $-25\%$ & $-67\%$ & $-48\%$ & $+0\%$ \\
\bottomrule
\end{tabular}
\end{table}

\begin{figure}[!htb]
\centering
\begin{tikzpicture}
\begin{groupplot}[group style={group size=3 by 1, horizontal sep=0.35cm, y descriptions at=edge left},
  width=0.39\linewidth, height=5.4cm, ymode=log, ymin=0.1, ymax=4.5,
  ytick={0.125,0.25,0.5,1,2,4}, yticklabels={0.125,0.25,0.5,1,2,4},
  xtick={0,...,5}, xticklabels={BE, NL, FR, DE-LU, DK1, DK2}, xmin=-0.6, xmax=5.6,
  x tick label style={font=\scriptsize, rotate=90, anchor=east}, y tick label style={font=\scriptsize},
  ymajorgrids, grid style={gray!20}, axis line style={gray!60}, tick style={gray!60},
  extra y ticks={1}, extra y tick labels={}, extra y tick style={grid style={black!50,dashed}},
  title style={font=\small, yshift=-1ex}]
\nextgroupplot[title={price}, ylabel={Rel.\ MAE (log scale)}, ylabel style={font=\scriptsize},
  legend to name=zoneleg, legend columns=6, legend style={font=\scriptsize, draw=none, /tikz/every even column/.append style={column sep=6pt}}]
\addplot[only marks, mark=square*, mark size=1.8pt, black!40] table[x expr={\thisrow{x}-0.30}, y=naivedaily] {zones_price.dat};
\addplot[only marks, mark=*, mark size=2pt, histc] table[x=x, y=chronos2] {zones_price.dat};
\addplot[only marks, mark=triangle*, mark size=2.4pt, orange!80!black] table[x expr={\thisrow{x}-0.06}, y=lgbmcov] {zones_price.dat};
\addplot[only marks, mark=*, mark size=2pt, covc] table[x expr={\thisrow{x}+0.06}, y=chronos2cov] {zones_price.dat};
\addlegendimage{only marks, mark=pentagon*, mark size=2.4pt, teal}
\addlegendimage{only marks, mark=diamond*, mark size=2.6pt, accent}
\legend{\m{naive\_daily}, \m{chronos2}, \m{lgbm\_cov}, \m{chronos2\_cov}, \m{tso\_cutoff}, \m{tso\_final}}
\nextgroupplot[title={wind and solar}]
\addplot[only marks, mark=square*, mark size=1.8pt, black!40] table[x expr={\thisrow{x}-0.30}, y=naivedaily] {zones_ws.dat};
\addplot[only marks, mark=*, mark size=2pt, histc] table[x=x, y=chronos2] {zones_ws.dat};
\addplot[only marks, mark=triangle*, mark size=2.4pt, orange!80!black] table[x expr={\thisrow{x}-0.06}, y=lgbmcov] {zones_ws.dat};
\addplot[only marks, mark=*, mark size=2pt, covc] table[x expr={\thisrow{x}+0.06}, y=chronos2cov] {zones_ws.dat};
\addplot[only marks, mark=pentagon*, mark size=2.4pt, teal] table[x expr={\thisrow{x}+0.18}, y=tsocutoff] {zones_ws.dat};
\addplot[only marks, mark=diamond*, mark size=2.6pt, accent] table[x expr={\thisrow{x}+0.30}, y=tsofinal] {zones_ws.dat};
\nextgroupplot[title={load}, xmax=3.6]
\addplot[only marks, mark=square*, mark size=1.8pt, black!40] table[x expr={\thisrow{x}-0.30}, y=naivedaily] {zones_load.dat};
\addplot[only marks, mark=*, mark size=2pt, histc] table[x=x, y=chronos2] {zones_load.dat};
\addplot[only marks, mark=triangle*, mark size=2.4pt, orange!80!black] table[x expr={\thisrow{x}-0.06}, y=lgbmcov] {zones_load.dat};
\addplot[only marks, mark=*, mark size=2pt, covc] table[x expr={\thisrow{x}+0.06}, y=chronos2cov] {zones_load.dat};
\addplot[only marks, mark=pentagon*, mark size=2.4pt, teal] table[x expr={\thisrow{x}+0.18}, y=tsocutoff] {zones_load.dat};
\end{groupplot}
\node at ($(group c2r1.north)+(0,1.0cm)$) {\pgfplotslegendfromname{zoneleg}};
\end{tikzpicture}
\caption{Pooled Rel.\ MAE per zone and target group (live, 2 October--2 November 2026, log scale; the dashed line is the
seasonal-naive baseline). Every point of a zone and panel is computed on the same races: those that all models shown
there forecast, the TSO references included where they exist (\m{tso\_cutoff}: BE and NL wind and solar, and load;
\m{tso\_final}: wind and solar outside Denmark, Section~\ref{sec:tso}). Danish load is left out (Section~\ref{sec:results}).}
\label{fig:zones}
\end{figure}
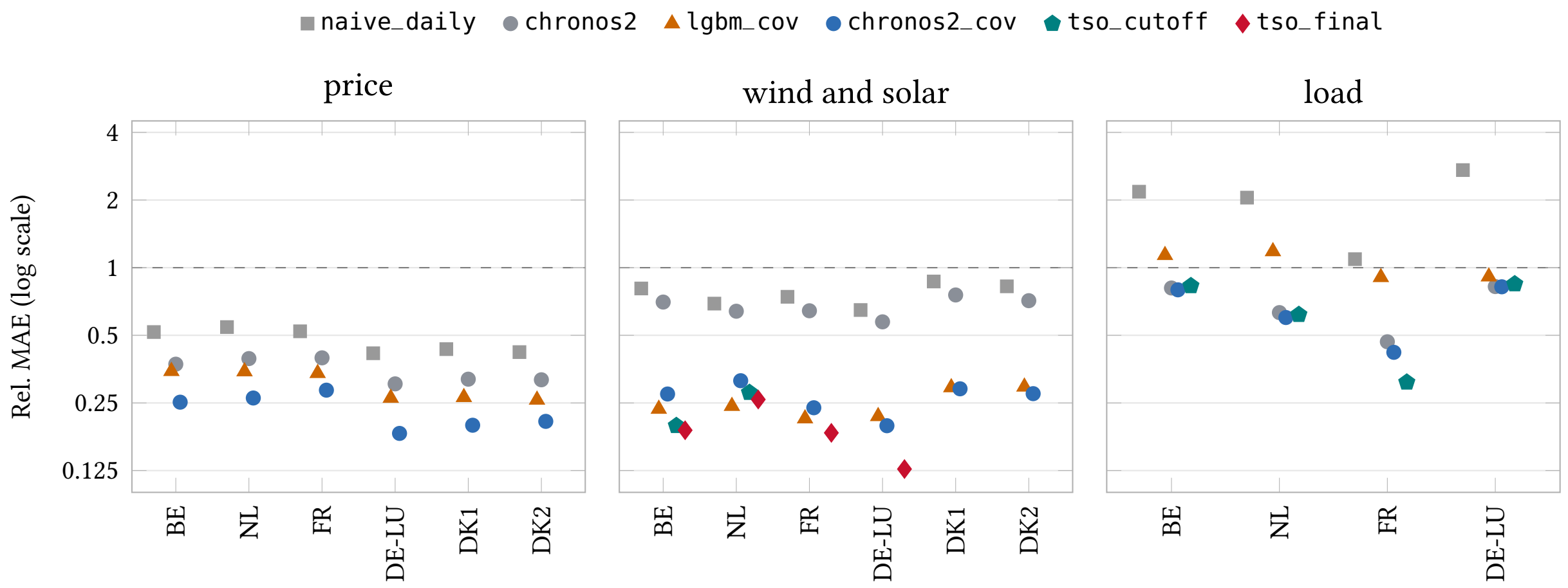

The effect does not depend on the zone (Figure~\ref{fig:zones}, where every point of a zone is computed on the same
races). In every zone the covariates reduce Chronos-2's error by 51--66\% on wind and solar and by 28--40\% on
price, while on load they reduce it by at most 10\%. On wind and solar the grid operators' forecasts
are ahead of Chronos-2 with covariates in every zone where they exist, and \m{lgbm\_cov} beats both of them in NL; on
load, Chronos-2 with covariates is ahead of the pre-gate forecast in BE, NL and DE-LU and behind it in FR.

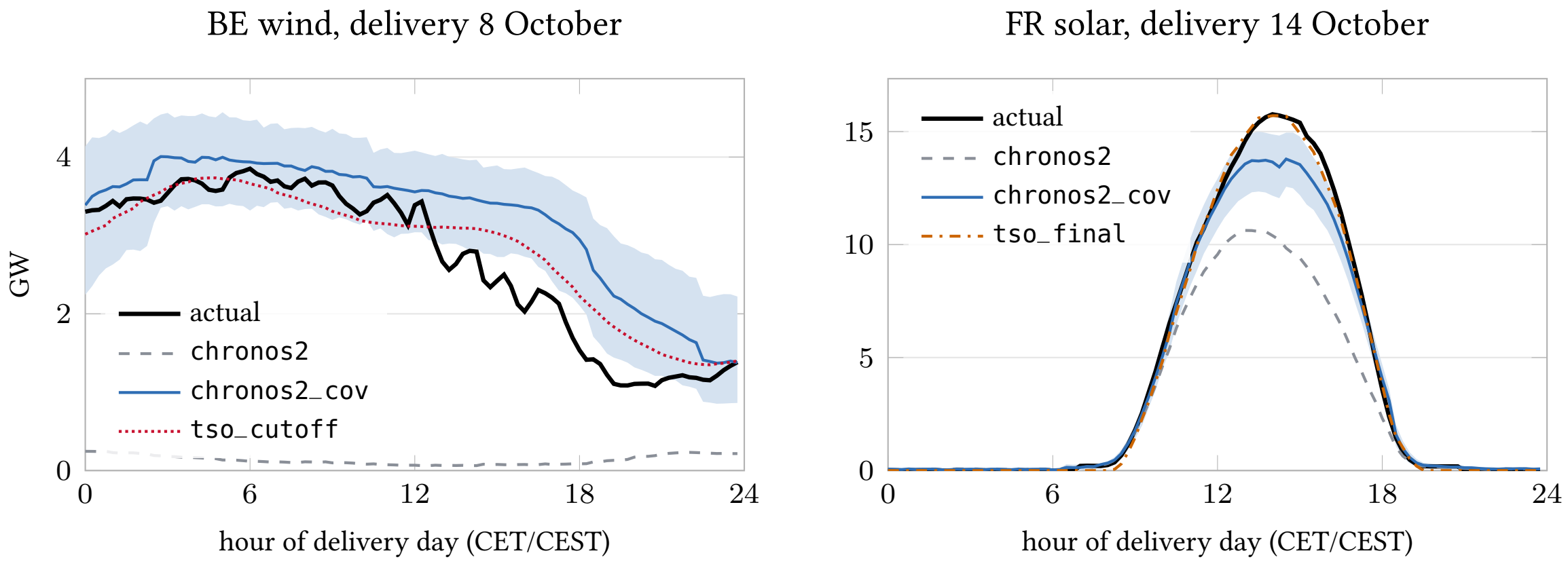
\begin{figure}[t]
\centering
\begin{tikzpicture}
\begin{groupplot}[group style={group size=2 by 1, horizontal sep=1.4cm},
  width=0.5\linewidth, height=5.4cm, xmin=0, xmax=24, xtick={0,6,12,18,24},
  xlabel={hour of delivery day (CET/CEST)}, xlabel style={font=\scriptsize},
  tick label style={font=\scriptsize}, ymajorgrids, grid style={gray!20}, axis line style={gray!60}, tick style={gray!60},
  legend style={font=\scriptsize, draw=none, fill=white, fill opacity=0.85, text opacity=1}, legend cell align=left]
\nextgroupplot[title={\small BE wind, delivery 8 October}, ylabel={GW}, ylabel style={font=\scriptsize}, ymin=0,
  ymax=5, legend pos=south west]
\addplot[name path=lo, draw=none, forget plot] table[x=h, y=lo] {example_BE_wind.dat};
\addplot[name path=hi, draw=none, forget plot] table[x=h, y=hi] {example_BE_wind.dat};
\addplot[covc!20, forget plot] fill between[of=lo and hi];
\addplot[black, very thick] table[x=h, y=actual] {example_BE_wind.dat};
\addplot[histc, thick, dashed] table[x=h, y=chronos2] {example_BE_wind.dat};
\addplot[covc, thick] table[x=h, y=chronos2_cov] {example_BE_wind.dat};
\addplot[accent, thick, densely dotted] table[x=h, y=tso_cutoff] {example_BE_wind.dat};
\legend{actual, \m{chronos2}, \m{chronos2\_cov}, \m{tso\_cutoff}}
\nextgroupplot[title={\small FR solar, delivery 14 October}, ymin=0, legend pos=north west]
\addplot[name path=lo, draw=none, forget plot] table[x=h, y=lo] {example_FR_solar.dat};
\addplot[name path=hi, draw=none, forget plot] table[x=h, y=hi] {example_FR_solar.dat};
\addplot[covc!20, forget plot] fill between[of=lo and hi];
\addplot[black, very thick] table[x=h, y=actual] {example_FR_solar.dat};
\addplot[histc, thick, dashed] table[x=h, y=chronos2] {example_FR_solar.dat};
\addplot[covc, thick] table[x=h, y=chronos2_cov] {example_FR_solar.dat};
\addplot[orange!80!black, thick, dashdotted] table[x=h, y=tso_final] {example_FR_solar.dat};
\legend{actual, \m{chronos2}, \m{chronos2\_cov}, \m{tso\_final}}
\end{groupplot}
\end{tikzpicture}
\caption{Two delivery days: actual output, Chronos-2 without and with covariates (shaded: the covariate version's
10--90\% band), and a TSO forecast (Elia's pre-gate forecast for BE; RTE's final forecast, published after the gate,
for FR).}
\label{fig:example}
\end{figure}

Figure~\ref{fig:example} illustrates the difference on two delivery days on which the covariates helped more than in
the median race. In the median wind or solar race, Chronos-2 with covariates has 41\% of the error of Chronos-2
without; on these two days the shares are 21\% and 28\%. On 8 October Belgian wind generation was between 3.1 and
3.9\,GW until midday and fell to about 1.1\,GW by the evening. Chronos-2 on its own history carried its recent output
forward and forecast less than 0.3\,GW for the whole day. The covariate version followed the forecast
weather, although, like Elia's pre-gate forecast, it anticipated the evening decline too late. On 14 October French solar peaked at 15.8\,GW: Chronos-2 alone reached
10.6\,GW, the covariate version 13.8\,GW, and RTE's final forecast, published after the gate, 15.7\,GW.

\subsection{Model size}

\begin{figure}[t]
\centering
\includegraphics[width=\linewidth]{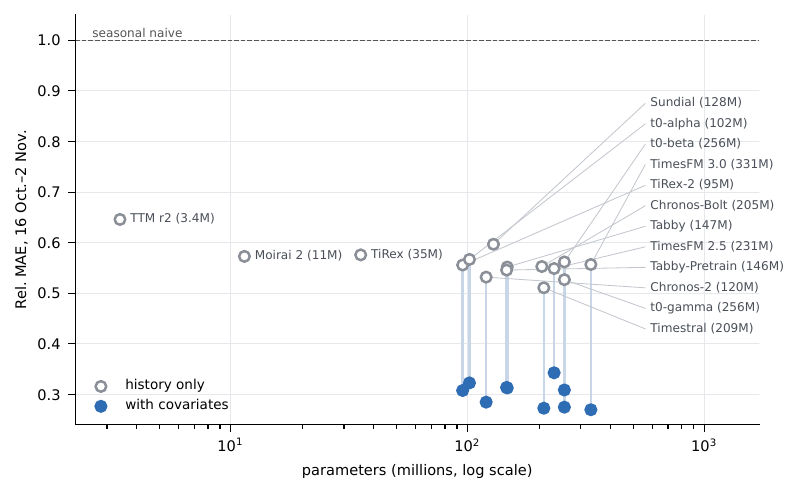}
\caption{Pooled Rel.\ MAE against parameter count for fifteen foundation models, without covariates (open) and,
where a variant exists, with the frozen covariates (filled, joined to the same model without), on the issue days
16 October--2 November 2026 (all zones, 390 races per model). Parameter counts are those of the checkpoints the benchmark loads;
TiRex-2 uses about 38M of its 95M in univariate mode.}
\label{fig:params}
\end{figure}

Figure~\ref{fig:params} sets each foundation model's error against its number of parameters, on the eighteen issue days
16 October--2 November, on which fifteen of them ran (390 races each; Toto left the field on 27 October, and Toto 2 and TimeH2O
joined after the 16th). Without covariates, larger models do somewhat better (Spearman $\rho = -0.55$, $p = 0.03$), but
the spread is small and not monotone: Timestral (209M), t0-gamma (256M) and Chronos-2 (120M) lead the rest by at least 0.01, TimesFM 3.0, the
largest at 331M, sits mid-field, and TTM r2 at 3.4M is the weakest. Above about 100M the relation disappears
($\rho = -0.18$, 11 models), and it is not significant with covariates ($\rho = -0.39$, ten models), where TimesFM 3.0,
Timestral and t0-gamma lead Chronos-2 by 0.01--0.02. Every covariate model, whatever its size, has a lower error than
every history-only model, so the weather input has a larger effect than any difference in size.

\subsection{Against purpose-built models}

The covariate-aware foundation models are used zero-shot, yet of the five that take covariates natively and have run
over almost the whole record, four have a lower error than a LightGBM model trained on the same covariate table, on the
races each shares with it (\m{lgbm\_cov}, Rel.\ MAE 0.330--0.337, against 0.282--0.320), and t0-alpha is level with it
(0.334 against 0.336). LightGBM receives more of the table: every NWP model's columns, the calendar and, for price,
yesterday's price, whereas the foundation models receive a small set of ensemble means and spreads. LightGBM is the
best of them on wind (0.152--0.154, against 0.160--0.196) and level on solar (0.379--0.396, against 0.336--0.398), but
considerably worse on load (1.018--1.026), where 28 days are too few to learn the response of load to the weather. Trained on
two years, \m{lgbm\_prod} does better. On the days both ran (16 October--2 November) it reaches 0.293, against 0.322 for
\m{lgbm\_cov}, within the range of the native-covariate foundation models (0.270--0.323). The direction agrees with FETS~\cite{fets}, where
covariate-informed zero-shot foundation models also outperformed gradient-boosted models trained on the target data,
although on fixed historical test blocks rather than live.

\subsection{Against the grid operators}
\label{sec:tso}

The grid operators' forecasts are compared with each entry on exactly the races both forecast. Two references are
used. The pre-gate forecast (\m{tso\_cutoff}) is the operators' forecast as published before the gate, and is therefore
information that a market participant could have used. The final forecast (\m{tso\_final}) is published between 16:15
and 18:00 on $D-1$, after the gate. It is based on later weather runs and more recent measurements than any entry can
use, and serves as an upper reference rather than as a competitor.

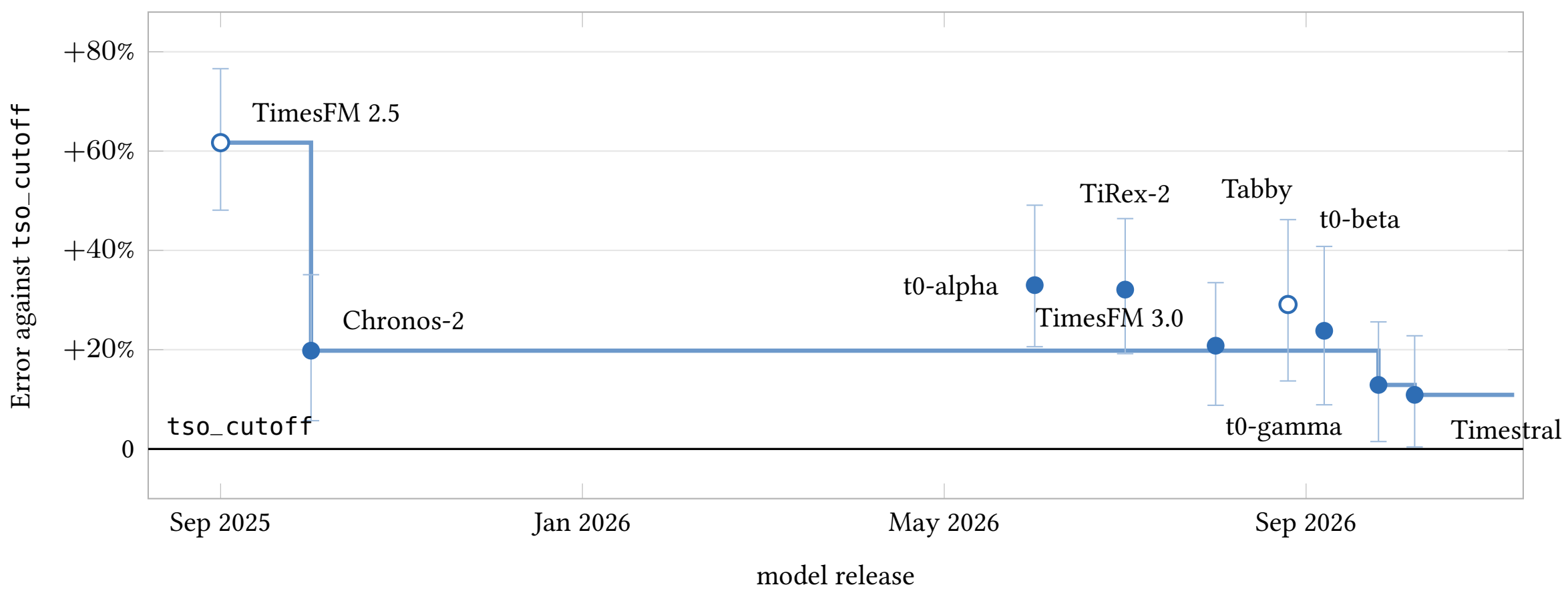
\begin{figure}[t]
\centering
\begin{tikzpicture}
\begin{axis}[
  width=0.95\linewidth, height=6.4cm, xmin=-0.8, xmax=14.4, ymin=-10, ymax=88,
  xtick={0,4,8,12}, xticklabels={Sep 2025,Jan 2026,May 2026,Sep 2026},
  x tick label style={font=\scriptsize}, y tick label style={font=\scriptsize},
  ytick={0,20,40,60,80}, yticklabels={0,$+20$\%,$+40$\%,$+60$\%,$+80$\%},
  xlabel={model release}, xlabel style={font=\scriptsize},
  ylabel={Error against \m{tso\_cutoff}}, ylabel style={font=\scriptsize},
  ymajorgrids, grid style={gray!20}, axis line style={gray!60}, tick style={gray!60}, clip=false]
\addplot[const plot, covc!70, very thick, forget plot] table[x=x, y=best] {release.dat};
\addplot[covc!70, very thick, forget plot] coordinates {(13.2,10.9) (14.3,10.9)};
\addplot[only marks, mark=*, mark size=2.3pt, covc, error bars/.cd, y dir=both, y explicit, error bar style={covc!45}]
  table[x=x, y=gap, y error plus=hi, y error minus=lo, restrict expr to domain={\thisrow{ridge}}{0:0}] {release.dat};
\addplot[only marks, mark=*, mark size=2.3pt, fill=white, draw=covc, thick, error bars/.cd, y dir=both, y explicit, error bar style={covc!45}]
  table[x=x, y=gap, y error plus=hi, y error minus=lo, restrict expr to domain={\thisrow{ridge}}{1:1}] {release.dat};
\draw[black, semithick] (axis cs:-0.8,0) -- (axis cs:14.4,0);
\node[anchor=south west, font=\scriptsize] at (axis cs:-0.75,0) {\m{tso\_cutoff}};
\begin{scope}[font=\scriptsize]
\node[anchor=south west] at (axis cs:0.2,63.2) {TimesFM 2.5};
\node[anchor=south west] at (axis cs:1.2,21.3) {Chronos-2};
\node[anchor=east] at (axis cs:8.75,32.5) {t0-alpha};
\node[anchor=south east] at (axis cs:10.8,21.8) {TimesFM 3.0};
\node[anchor=south east, inner xsep=1pt] at (axis cs:11.9,46.6) {Tabby};
\node[anchor=east] at (axis cs:12.55,4.0) {t0-gamma};
\node[anchor=west] at (axis cs:13.45,4.0) {Timestral};
\node[anchor=south west, inner xsep=1pt] at (axis cs:12.1,41.8) {t0-beta};
\node[anchor=south] at (axis cs:10.0,46.9) {TiRex-2};
\end{scope}
\end{axis}
\end{tikzpicture}
\caption{Foundation models with covariates against the grid operators' pre-gate forecast, by model release date, with
day-bootstrap 95\% intervals; the line is the best model released so far. Open markers: covariates through a ridge
regression. Geometric mean of the per-series error ratios on the
same 141 races (issue days 16 October--2 November 2026, eight series); every model ran on the same days, so the figure compares
model generations, not progress over time.}
\label{fig:release}
\end{figure}

Table~\ref{tab:tsogap} and Figure~\ref{fig:release} set the foundation models with covariates against \m{tso\_cutoff} on the same 141 races: the
eighteen issue days 16 October--2 November, BE and NL wind and solar, and load in BE, NL, FR and DE-LU. The operators remain
ahead overall. The closest single models, Timestral and t0-gamma, have 11\% and 13\% more error (95\% day-bootstrap
intervals $0$ to $+23$ and $+1$ to $+26$). TimesFM 2.5, which takes the covariates through a ridge regression and gains
least from them (Table~\ref{tab:cov}), has 62\% more. Within a model family the newer release has the lower error on
these races: TimesFM 3.0 has 25\% less error than TimesFM 2.5 (interval 21--29) and t0-gamma 15\% less than t0-alpha
(9--21).

Figure~\ref{fig:release} places the same gaps against each model's release date, with the best model released so far
as a frontier. The frontier falls from $+62$\% (TimesFM 2.5, September 2025) to $+20$\% a month later (Chronos-2) and
to $+11$\% (Timestral, October 2026). Only the first step is larger than the intervals; the later releases improve on
Chronos-2 by amounts within them. All models ran over the same weeks, so the figure compares model generations on a
common test, not the progress of one forecaster over time. TimesFM 2.5 and Tabby take the covariates through a ridge
regression, which may explain part of their distance.

\begin{table}[t]
\centering\small
\caption{Foundation models with covariates and \m{lgbm\_prod} against the pre-gate TSO forecast \m{tso\_cutoff}, on the
same 141 races (issue days 16 October--2 November 2026; BE and NL wind and solar, load in BE, NL, FR and DE-LU): extra pooled
error, geometric mean over series, for all eight series with its day-bootstrap 95\% interval and per target group.
Negative: the entry has the lower error.}
\label{tab:tsogap}
\begin{tabular}{@{}lrcrr@{}}
\toprule
Model & All & 95\% interval & Wind, solar & Load \\
\midrule
Timestral & $+11\%$ & [$0$, $+23$] & $+16\%$ & $+6\%$ \\
t0-gamma & $+13\%$ & [$+1$, $+26$] & $+22\%$ & $+4\%$ \\
Chronos-2 & $+20\%$ & [$+6$, $+35$] & $+32\%$ & $+8\%$ \\
TimesFM 3.0 & $+21\%$ & [$+9$, $+33$] & $+21\%$ & $+21\%$ \\
t0-beta & $+24\%$ & [$+9$, $+41$] & $+34\%$ & $+15\%$ \\
Tabby & $+29\%$ & [$+14$, $+46$] & $+48\%$ & $+13\%$ \\
TiRex-2 & $+32\%$ & [$+19$, $+46$] & $+31\%$ & $+33\%$ \\
t0-alpha & $+33\%$ & [$+21$, $+49$] & $+35\%$ & $+31\%$ \\
TimesFM 2.5 & $+62\%$ & [$+48$, $+77$] & $+52\%$ & $+72\%$ \\
\midrule
\m{lgbm\_prod} & $+11\%$ & [$+2$, $+20$] & $+1\%$ & $+22\%$ \\
\bottomrule
\end{tabular}
\end{table}

The gap differs by target. On load the closest models have slightly more error than the operators overall
(Timestral $+6$\%, t0-gamma $+4$\%, with intervals that include zero): they have 1--11\% less error in BE, NL and
DE-LU, and 41--59\% more in FR. On wind and solar they have 16--22\% more error. There \m{lgbm\_prod}, trained on two
years of these series, is level with the operators ($+1$\%, interval $-8$ to $+11$), but it has 22\% more error on
load. Over the fourteen days on which the ensembles ran (109 races), \m{ens\_hybrid} has 5\% more error than the
operators overall (interval $-6$ to $+18$), 10\% more on wind and solar and the same on load. On wind and solar most of the improvement over a history-only model comes from the weather
input. On the races of Table~\ref{tab:tsogap}, TimesFM 3.0 reaches 0.61 of the baseline's error on wind from its history
alone and 0.17 with covariates, against 0.14 for the operators; on solar the values are 0.94 and 0.40, against 0.34. On
load it does not change (0.68), against 0.56. Outside NL the pre-gate load reference is the ENTSO-E load
forecast, which the load models received as an input until 15 October.

The final forecasts, published after the gate, are further ahead. On wind and solar in BE, NL, FR and DE-LU (98 races
each) they reach Rel.\ MAE 0.118 and 0.285, against 0.143 for \m{lgbm\_cov} on wind and 0.333 for TimesFM 3.0 on solar,
the best entries on those races. Over its 18 days \m{lgbm\_prod} comes closest on wind (0.128, against 0.117). Danish
wind is left out of this comparison because Energinet's final forecast and the actuals do not measure the same
quantity: its day-ahead forecast for DK1 wind has run at about 0.8 of measured output for two years, a difference in
level rather than in skill.

\section{Limitations}
\label{sec:limits}

The record is short and covers one season. Electricity series are not stationary: autumn has neither the solar peaks of summer nor
the demand peaks of winter, and installed capacity and market rules
change from year to year. Which models lead may differ between such regimes, so the standings are provisional and
will change as the record grows. The three zones added on 15 October have nineteen issue days, and entries that
joined late are compared with the rest only on the days they share.

The weather inputs are three deterministic models from three centres (DWD, NOAA and ECMWF), and their spread stands
in for forecast uncertainty. All three are served by one provider, Open-Meteo, which is a single point of failure: on
six of the thirty-two live days it served an ICON-EU run before the run was fully ingested.\footnote{On 3, 4, 5, 10, 13
and 17 October the 06 UTC ICON-EU run was served at the cut-off with missing values over the delivery day. Models
built on the three-model ensemble means were barely affected; models that read each NWP model's own columns lost
accuracy on those days. A completeness check has rejected such runs since 19 October; the affected days are scored as
they were run.} The two-year training frames match the lead time of the live runs but take ICON-EU and GFS from a
previous-runs product that can contain runs initialised after the cut-off of the day they describe; this concerns
the training data of two entries, never what is scored.

A fair evaluation depends on details that are rarely examined: the formal definition of each actual and of each
operator forecast, and the way the published data are read. The record has several examples, each documented and,
where the forecast could not be repaired, voided. On seven issue days in DK1 and three in DK2, Energinet's load
forecast on ENTSO-E ended at about 11:00 on the delivery day when the run read it, with a closing zero that was parsed
as zero load for the rest of the day. Those \m{tso\_cutoff} forecasts are voided, since the later hours, published
after the gate, cannot repair them. Until the parser was fixed on 31 October the same zeros reached the load forecast
input of the price models; those forecasts are scored as they were run (Danish price errors show no visible harm on
those days, but the counterfactual was not run). The Dutch targets changed definition during the record
(Section~\ref{sec:actuals}), the Danish final forecasts measure a different scope from the actuals
(Section~\ref{sec:tso}), the \m{tso\_cutoff} reference exists only where a TSO publishes before the gate (BE and NL
wind and solar, and load), and the weather points rest on public capacity registers of uneven detail (municipal data
in BE, NL and DK, commune centres in FR, unit positions in DE-LU). A shared, explicit definition of each target, its
source and its scope, recorded with the scores, makes such differences visible before they reach a ranking.

Each entry's configuration (its inputs, context length and settings) was fixed before it went live, so no entry was
tuned on the days on which it is scored; the configurations were, however, chosen by authors who had seen the data of
the preceding weeks. The covariate models are deliberately simple and untuned; \m{lgbm\_cov} and \m{lin\_cov} train
on the 28-day context for comparability with the foundation models, a known handicap that \m{lgbm\_prod} removes. A
model that fails is left out of the races it missed (Section~\ref{sec:ranking}), so a model that fails often is
flagged by the run tracker rather than penalised in the rating.

Pooled errors and the rating answer different questions. A pooled error, even scaled by the seasonal-naive baseline,
depends on the days it is pooled over, since some days are harder than others for every model, so two entries can be
compared on it only over the same days: Timestral without covariates has a pooled error of 0.511 over its eighteen
days and of 0.443 over the last nine, on which Toto 2 ran. The rating compares entries within each race, where the difficulty of the day is shared,
and connects entries that ran on different days through the opponents they have in common, so it places the whole
field on one scale, with wider intervals for entries with fewer races. The rating counts wins, not their margins. A model that beats another by a wide margin gains as much from the game as
one that wins narrowly, and the rating does not compare the margin a model achieves with the margin expected from the
two ratings. The pooled relative errors, reported next to the rating, carry the margin: on wind and solar \m{lgbm\_prod}
has the lower error than the operators' pre-gate forecast in only 44\% of their races, while its pooled error is level
with theirs (Section~\ref{sec:tso}). Rating systems in sports, from which the race format is borrowed, have been extended to
margins of victory, either by scaling each update with the margin, as in FiveThirtyEight's Elo
ratings~\cite{fivethirtyeight}, or by modelling the margin itself~\cite{kovalchik,szczecinski}; least-squares ratings
of the Massey type~\cite{massey} fit the margins directly. A rating of this kind on the relative errors should be
considered in later versions.

The forecasts are fixed before the outcome is known, but the analyses in Section~\ref{sec:results} were not
registered in advance. Their windows follow from the dates on which entries joined, and the authors chose the
comparisons with the record in view. This is the third problem of the introduction, in reduced form: the scores,
forecasts and analysis scripts are public, so each comparison can be repeated on other windows, but only a set of
comparisons fixed before the data arrive would remove the freedom entirely. Finally, the results describe only the
entries in the field.

\section{Conclusion}

A live benchmark aligned with the market gate excludes look-ahead and fixes the evaluation days before the outcomes
are known. EnergyBench runs such a benchmark daily for the European day-ahead market, on quarter-hours, from frozen
public data with recorded provenance, with rankings reported with intervals. Over the first thirty-two live days in six zones, the weather information available at the cut-off
mattered more than the choice of model. The same foundation model forecasts with about half the error when it is given that
information (for Chronos-2, in every zone). Zero-shot foundation models with covariates match or outperform a
gradient-boosting model trained on the same covariate table, and the median of several such models has a lower error
than any of them. Model size has a much smaller effect. The grid operators' pre-gate forecasts remain ahead on wind and solar,
where the best single foundation
models have 16--22\% more error, while on load the gap is small (4--6\%); the operators'
final forecasts, published after the gate, are further ahead. Further zones, references and settlement rules are
planned.

\end{document}